\documentclass{article}

\usepackage[preprint]{neurips_2026}

\usepackage[utf8]{inputenc} 
\usepackage[T1]{fontenc}    
\usepackage{hyperref}       
\usepackage{url}            
\usepackage{booktabs}       
\usepackage{amsfonts}       
\usepackage{nicefrac}       
\usepackage{microtype}      
\usepackage{xcolor}         
\usepackage{amsmath}
\usepackage{amssymb}
\usepackage{mathtools}
\usepackage{amsthm}
\usepackage{graphicx}
\usepackage{subcaption}
\usepackage{booktabs} 
\usepackage{url}
\usepackage{array}
\usepackage{placeins}
\usepackage[most]{tcolorbox}
\usepackage{soul}
\usepackage{xcolor} 
\usepackage{wrapfig}
\usepackage{mdframed}
\usepackage{enumitem}
\usepackage{setspace}
\usepackage{caption}
\usepackage{adjustbox}
\usepackage{multirow}
\usepackage{algorithm}
\usepackage{algpseudocode}

\title{EQPO: Equitable Group Relative Policy Optimization for Clinical Reasoning}

\author{%
  Shiqi Dai$^{1}$, Wei Dai$^{1}$, Jiaee Cheong$^{2}$, Paul Pu Liang$^{1}$ \\
  $^{1}$ MIT, $^{2}$ Harvard University
}

\begin{document}

\maketitle

\begin{abstract}
Medical AI systems demonstrated impressive diagnostic performance, yet they routinely show uneven accuracy across demographic groups, disadvantaging underrepresented populations. Although multimodal reasoning foundation models have pushed clinical diagnosis forward, reinforcement learning-based post-training tends to absorb and magnify the biases present in majority-dominated training corpora. We propose \textbf{Equitable Group Relative Policy Optimization (EQPO)}, a hierarchical reinforcement learning method that encourages balanced learning across heterogeneous clinical populations by adaptively reweighting samples according to subgroup representation, task difficulty, and data source. As demographic annotations are frequently missing in real-world clinical data, EQPO additionally applies unsupervised clustering to recover latent subpopulations when they are unavailable. On 7 diagnostic benchmarks covering 5 modalities (X-ray, CT, dermoscopy, mammography, ultrasound), EQPO reduces F1 standard deviation by 43.9\% and the maximum cross-group F1 gap by 42.7\% on QoQ-Med3-8B over vanilla GRPO, and narrows predictive parity gaps by 27.2\% on MedGemma-4B over bias-mitigated RL baselines while raising F1 by 12.5\% even without any demographic labels. Examining the training trajectory shows that EQPO steadily improves fairness over the course of optimization, in contrast to baseline methods whose fairness degrades as training proceeds, and the discovered implicit groups remain stable and align with masked demographic attributes. We further release \textbf{EquiMedGemma-4B} and \textbf{EquiQoQ-Med3-8B}, equitability-aware clinical VLLMs that attain state-of-the-art accuracy with markedly smaller demographic gaps. Code, models, and the evaluation framework are available at \href{https://anonymous.4open.science/r/fairness_submission-D923/}{this anonymous link}.
\end{abstract}

\section{Introduction}

Medical artificial intelligence (AI) has demonstrated strong capabilities in processing vast amounts of clinical data with both accuracy and efficiency~\citep{rajpurkar2022ai, shuja2024harnessing}. These systems have shown particular promise in detecting subtle health indicators that may escape human observation, substantially enhancing diagnostic precision while reducing healthcare costs~\citep{dai2025developing, sun2022machine}. Recent advances in multimodal large language models (MLLMs) have further expanded these capabilities, enabling integrated analysis across diverse clinical modalities including imaging, time series, and textual records~\citep{cui2024biomedicalvital, dai2025qoq, zhang2024generalist,zhu2024uni}. Beyond pattern recognition, these models can articulate their diagnostic reasoning in natural language \cite{amann2020explainability}, offering clinicians interpretable rationales that support shared decision-making. 

However,
medical AI systems performing clinical reasoning can exhibit troubling performance disparities across demographic subpopulations, undermining their equitable deployment in practice. Studies have revealed that clinical datasets are overwhelmingly skewed toward majority groups, whether defined by race, gender, age, or socioeconomic status~\citep{larrazabal2020gender, obermeyer2019dissecting, liang2021towards, thakur2023language}, leading state-of-the-art classifiers to demonstrate significant true positive rate disparities across clinical tasks and demographic subgroups~\citep{seyyedkalantari2021chexclusion, seyyedkalantari2021underdiagnosis}. During training, conventional optimization naturally favors well-represented populations, as they contribute more gradient updates and dominate the loss landscape~\citep{stiglic2020interpretability, kumarakulasinghe2020evaluating}. The heterogeneous nature of multimodal clinical data, spanning multiple specialties and patient demographics, can further exacerbate these disparities as different groups may require fundamentally different diagnostic considerations~\citep{ghanvatkar2023graph, cui2023deep}. Such systematic biases not only perpetuate healthcare inequalities but also erode trust in AI-assisted diagnosis, particularly among underserved communities who stand to benefit most from improved healthcare access~\citep{trustinai}.

Current approaches to mitigating bias in medical AI typically rely on data augmentation, reweighting schemes, or post-hoc calibration~\citep{teng2022survey,khan2023fair,mehta2024evaluating}. While these have been studied for discriminative models~\citep{lahoti2020fairness}, their application to \textit{generative clinical reasoning}, where \textit{models produce free-form diagnostic explanations rather than fixed-label predictions}, presents distinct challenges that prior work does not address. Equitability-aware optimization techniques like group distributionally robust optimization (DRO)~\citep{sagawa2019distributionally} were also designed for prediction models and cannot be directly applied to the open-ended reasoning of modern LLMs. Furthermore, while reinforcement learning has improved LLM alignment for helpfulness and harmlessness~\citep{ouyang2022training, bai2022training}, its application to equitability in medical reasoning remains unexplored. Equitability in medical settings is particularly challenging given that diagnosis relies on comprehensive multi-symptom analysis, data availability varies across domains, collection is skewed toward those with healthcare access, and demographic annotations are frequently unavailable due to privacy regulations or incomplete records.

To close this gap, we introduce \textbf{Equitable Group Relative Policy Optimization (EQPO)}, a hierarchical RL approach that promotes equitable learning across heterogeneous clinical populations. Unlike classical importance weighting methods that operate on fixed loss functions, EQPO addresses the unique challenges of policy optimization by integrating equitability constraints directly into the advantage estimation process while preserving the convergence properties of conventional GRPO. For settings where demographic labels are unavailable, we introduce a reward-based clustering approach that discovers latent groups based on task-specific difficulty patterns rather than extracted features, enabling the method to identify and upweight samples that the model finds challenging, which empirically correlate with underrepresented demographics. In summary, our work makes several primary contributions:

\begin{enumerate}[noitemsep,topsep=0pt,nosep,leftmargin=*,parsep=0pt,partopsep=0pt]
    \item We propose \textbf{EQPO}, one of the first fair RL algorithms for clinical reasoning models, employing \textit{adaptive importance weighting} based on demographic representation and task difficulty. Our method inherits GRPO's convergence properties through batch-wise advantage renormalization while incorporating self-regulating scaling that equilibrates as minority group performance improves. Empirically, on QoQ-Med3-8B, EQPO reduces F1 standard deviation by 43.9\% and the maximum cross-group F1 gap by 42.7\% over vanilla GRPO, and on MedGemma-4B it cuts predictive parity by 27.2\% over bias-mitigated baselines while raising F1 by 12.5\%. Training dynamics also show progressive equitability improvement rather than the deterioration observed in baseline RL methods.
    
    \item We demonstrate that EQPO can implicitly discover underrepresented groups without demographic labels through reward-based clustering, achieving positive correlation with ground-truth demographics. The learned scaling factors systematically upweight minority samples, and the discovered clusters remain stable throughout training.
    
    \item We release \textbf{EquiMedGemma-4B} and \textbf{EquiQoQ-Med3-8B}, equitability-aware clinical VLLMs trained with EQPO across 7 datasets spanning 5 modalities. Both attain SOTA performance while demonstrating significantly reduced demographic disparities, representing the first publicly available clinical VLLMs explicitly optimized for equitability through RL.
\end{enumerate}

Finally, we publicly release our models, training pipeline, and comprehensive equitability evaluation metrics to facilitate reproducible research in equitable medical AI. By addressing equitability as a fundamental optimization objective rather than a post-hoc consideration, our work establishes a new paradigm for developing clinical AI systems that serve all populations equitably.
\section{Related Work}

\noindent\textbf{Fairness in Clinical AI and Large Language Models.}
While unimodal clinical diagnosis leverages single data sources such as images \citep{khan2023fair,mehta2024evaluating} or tabular data \citep{dehghani2024fairness,roosli2022peeking}, multimodal methods fuse multiple modalities to learn richer representations, consistently outperforming unimodal approaches \citep{liang2024foundations,daiclimb,alsaad2024multimodal}. The increasing adoption of foundation models in healthcare \citep{daiclimb,jin2024fairmedfm,luo2024fairclip} amplifies fairness challenges, as integrating multiple knowledge sources can exacerbate biases across fused modalities, and recent work documents fundamental limits of post-hoc LLM debiasing \citep{anthis2025impossibility, bender2021dangers, weidinger2021ethical}. Recent multimodal LLMs such as Qwen-2.5-VL \citep{bai2025qwen2} and domain-specific models like MedGemma \citep{sellergren2025medgemma} have demonstrated impressive clinical reasoning capabilities, yet their fairness properties remain largely unexplored. Existing fairness works in healthcare \citep{khan2023fair,luo2024fairclip} have focused on predictive bias in unimodal models for single clinical tasks, such as chest radiograph analysis \citep{khan2023fair} or glaucoma detection \citep{luo2024fairclip}. Our work presents the first attempt to evaluate and optimize fairness on a VLLM trained across multiple clinical tasks and domains simultaneously.

\noindent\textbf{Fairness in Reinforcement Learning.}
Reinforcement learning methods which typically attempt to maximize agent reward may neglect fairness considerations \citep{jabbari2017fairness,smith2023bias}; prior fairness-aware RL work targets recommendation \citep{ge2022toward} and RLHF preference aggregation \citep{chakraborty2024maxmin}, but in single-task or value-based regimes that do not transfer to critic-free generative VLLMs. Recent critic-free RL algorithms for LLMs, such as GRPO \citep{shao2024deepseekmath}, RLOO \citep{rloo}, and REINFORCE++ \citep{hu2025reinforce}, have demonstrated success in aligning language models without value function estimation, but lack mechanisms to address fairness across heterogeneous populations. Traditional fairness mitigation methods include resampling \cite{puyol2021fairness} and Group DRO \cite{sagawa2019distributionally}, though these were designed for discriminative models with fixed output spaces. Adversarially Reweighted Learning (ARL) \citep{lahoti2020fairness} addresses fairness without demographics, but requires training a separate adversary network, assumes fixed output spaces inherent to discriminative classifiers, and cannot handle the generative, multi-step reasoning of VLLMs. To the best of our knowledge, no current work addresses fairness in critic-free RL optimization of VLLMs, where computational requirements and multi-step reasoning processes present unique challenges distinct from traditional RL settings. Our work bridges this gap by extending GRPO with fairness-aware mechanisms specifically designed for medical VLLMs.
\section{The EQPO Method}
\label{sec:eqpo}

\begin{figure*}
    \centering
    \vspace{-4mm}
    \includegraphics[width=\linewidth]{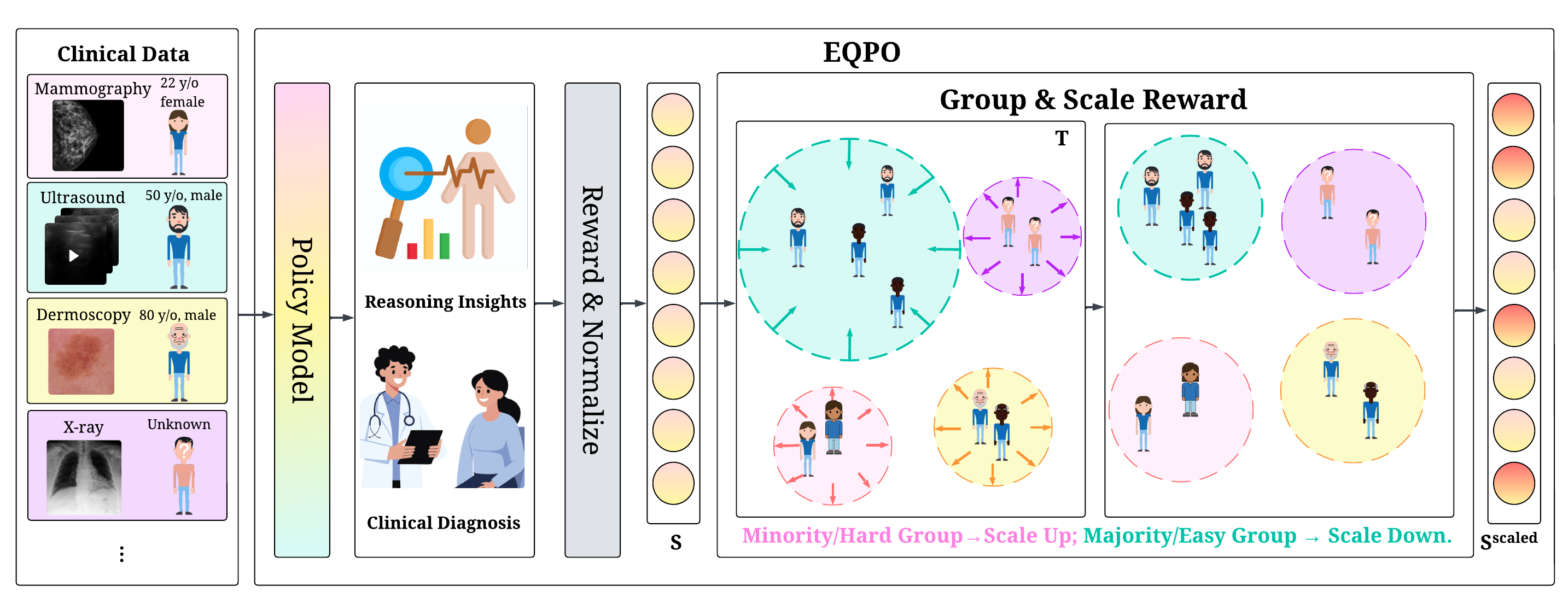}
    \caption{\textbf{EQPO Training Pipeline.} Our method addresses equitability disparities by adaptively scaling rewards based on demographic representation and task difficulty. Starting with medical data containing both labeled and unlabeled samples, the policy model generates multiple responses for each prompt, producing both reasoning insights and clinical diagnoses. These responses are evaluated and assigned rewards. EQPO then groups the rewards by explicit demographic groups where available. For samples with unavailable demographic information, we employ K-means clustering to discover implicit groups. Then, \textcolor[HTML]{ff80df}{minority or challenging groups} receive amplified learning signals through inverse temperature scaling, while \textcolor[HTML]{00c2a8}{majority or well-represented groups} are scaled down. This ensures that the model learns equitably from all subpopulations, preventing the typical bias toward majority groups that occurs in standard training. }
    \label{fig:eqpo}
    \vspace{-1.3em}
\end{figure*}

Medical AI systems often exhibit performance disparities across demographic subpopulations, reflecting biases inherent in training data distributions \citep{luo2024fairclip,khan2023fair}. While Group Relative Policy Optimization (GRPO) has demonstrated success in language model alignment through within-group reward normalization, it lacks mechanisms to address systematic sub-group imbalances across heterogeneous populations and instead tends to inherit, and often amplify, the demographic skew of its training data. We introduce EQPO, a hierarchical scaling approach that promotes equitable learning by adaptively weighting contributions from different domains and demographic groups based on their demographic information and difficulty measured via model performance. EQPO enables us to train EquiMedGemma-4B and EquiQoQ-Med3-8B, state-of-the-art models balancing clinical reasoning with equitability across populations. A pseudocode overview of the method is provided in App.~\ref{app:pseudocode}.

\textbf{Background: Group Relative Policy Optimization (GRPO).} GRPO operates by normalizing rewards within groups of responses to identical prompts, eliminating the need for value function estimation. For a prompt $q$ generating response group $G_{(q,t)}$ at iteration $t$, each response $o_{(q,i,t)}$ receives reward $r_{(q,i,t)}$. The advantage is computed as $\hat{A}_{(q,i,t)}^{\text{GRPO}} = \frac{r_{(q,i,t)} - \hat{\mu}_{G_{(q,t)}}}{\hat{\sigma}_{G_{(q,t)}} + \varepsilon}$, ensuring zero mean and unit variance within each response group. This normalization enables fair comparison among responses to the same prompt but treats all prompts equally, regardless of their source domain or demographic representation.

\textbf{The Fairness Challenge.} Consider a training dataset where prompts originate from different domains $g \in \mathcal{G}$ and are associated with demographic groups $d \in \mathcal{D}_{\text{demo}}$. Each prompt $q$ at iteration $t$ belongs to exactly one domain $g_{(q,t)}$ and one demographic group $d_{(q,t)}$.

Standard GRPO optimization naturally favors well-represented domain-demographic pairs as they contribute more gradient updates. Furthermore, normalizing advantages by within-prompt reward standard deviation systematically upweights low-variance prompts \citep{liu2025understanding}, which in clinical data tend to correspond to the easy, well-represented majority cases that the model already handles. When all rollouts for a prompt agree, GRPO yields zero advantage and therefore no gradient signal \citep{yu2025dapo}, silencing precisely the hard minority cases on which the policy uniformly fails. GRPO is also prone to entropy collapse during long training runs \citep{cui2025entropy}, locking the policy onto whichever distribution dominates each batch. Together these effects create a feedback loop where the model becomes increasingly specialized for majority populations while performance on minority groups stagnates. EQPO breaks this cycle through adaptive importance weighting that inversely correlates with group representation and performance; we formalize this argument and prove a strict reduction in cross-group accuracy variance under a bounded KL budget in App.~\ref{app:theory}.

\textbf{Hierarchical Scaling Framework.} EQPO implements a three-stage process that transforms GRPO's uniform treatment into demographically aware optimization:

\textit{(i) Normalization:} We first apply standard GRPO normalization to obtain $s_{(q,i,t)} = \frac{r_{(q,i,t)} - \hat{\mu}_{G_{(q,t)}}}{\hat{\sigma}_{G_{(q,t)}} + \varepsilon}$. This standardizes each response's reward relative to the mean $\hat{\mu}_{G_{(q,t)}}$ and standard deviation $\hat{\sigma}_{G_{(q,t)}}$ of its rollout group.

\textit{(ii) Group Discovery:} In medical datasets, demographic labels may be incomplete or unavailable for certain samples. We define \textit{explicit groups} as those with \textit{labeled} demographic attributes such as age or gender, while \textit{implicit groups} are latent subpopulations discovered through unsupervised clustering when such labels are missing. To identify implicit groups, we leverage the model's performance patterns: within each domain $g$, we construct feature vectors $\mathbf{v}_q \in \mathbb{R}^{|G_{(q,t)}|}$ for each unlabeled prompt $q$, where each dimension represents the raw reward from a different rollout. In GRPO, a rollout refers to a single generated response for a given prompt, with multiple rollouts per prompt enabling reward normalization across response variations. For instance, a chest X-ray prompt without demographic labels might generate 5 rollouts with rewards [0.2, 0.8, 0.7, 0.9, 0.3], forming its feature vector.

This reward-based representation offers two key advantages over traditional feature extraction methods. First, it provides computational efficiency, requiring only a vector of length equal to the number of rollouts rather than high-dimensional CNN or ViT embeddings. More importantly, it \textit{directly captures task-specific difficulty patterns rather than input-level similarities}. While visual features might group images by appearance, our approach groups samples by their inherent diagnostic challenge, ensuring that cases with similar learning difficulties receive similar treatment regardless of visual characteristics. K-means clustering then groups prompts with similar reward distributions, where well-represented cases typically form larger clusters with higher rewards, while challenging cases form smaller clusters with lower or more variable rewards. The optimal number of clusters is determined via the elbow method \citep{thorndike1953belongs}. Crucially, because our scaling mechanism inversely weights rewards by cluster size and performance as shown in Eq.~\ref{temp1}, smaller clusters representing rarer cases receive amplified learning signals, as evidenced in Fig.~\ref{fig:per_group_analysis}(a), ensuring that unlabeled minority subpopulations benefit from equitability-aware optimization.


\textit{(iii) Demographic-Group-Based Reward Scaling:} We compute hierarchical temperature factors that capture both representation and difficulty. 
At the domain and group level, this is represented by:
\begin{equation}
T_{(g,t)} = \sqrt{N_{(g,t)}} \cdot \bar{r}_{(g,t)},T_{(\gamma,g,t)} = \sqrt{N_{(\gamma,g,t)}} \cdot \bar{r}_{(\gamma,g,t)}.
\label{temp1}
\end{equation}
respectively for group $\gamma$ (explicit or implicit) in domain $g$. $N_{(g,t)}$ counts samples in domain $g$ and $\bar{r}_{(g,t)}$ represents the domain's mean raw reward. The normalized rewards undergo inverse temperature scaling:
\begin{equation}
s_{(q,i,t)}^{\text{scaled}} = \frac{s_{(q,i,t)}}{\max(T_{(g_{(q,t)},t)} \cdot T_{(\gamma_{(q,t)},g_{(q,t)},t)}, \varepsilon)},
\end{equation}
thus amplifying signals from underrepresented or challenging groups while attenuating those from dominant populations. Lastly, following \citep{schulman2017proximal}, we renormalize the advantage to zero mean and unit variance with $\hat{A}_{(q,i,t)}^{\text{EQPO}} = \frac{s_{(q,i,t)}^{\text{scaled}}}{\sigma_{\text{batch}}}$, where $\sigma_{\text{batch}}$ denotes the standard deviation across all scaled rewards in current batch.

\textbf{Training Objective.} EQPO retains GRPO's policy gradient formulation with clipped importance sampling:
\vspace{-0.5em}
\begin{equation}
    J_{\text{EQPO}}(\theta) = \mathbb{E}_{q,o}\left[ \sum_{k=1}^{n_o} \min\Big(\varphi_k(\theta) \hat{A}^{\text{EQPO}},\text{clip}(\varphi_k(\theta), 1\pm\varepsilon) \hat{A}^{\text{EQPO}}\Big) - \beta D_{\text{KL}}(\pi_\theta \| \pi_{\text{ref}}) \right].
\end{equation}
where $\varphi_k(\theta)$ represents the importance ratio at token $k$, and the advantage now incorporates equitability-aware scaling.

\textbf{Convergence Analysis.} We provide a sketch of the method convergence here and give the full derivation in App.~\ref{app:theory}. In summary, \textit{we prove that under a bounded KL budget, EQPO converges to a strictly lower cross-group accuracy variance $\sigma_{\text{Acc}}$ than GRPO}. Consider two groups $a, b$ with $N_a \ll N_b$ and approximately equal initial rewards $\bar{r}_a \approx \bar{r}_b$. Decomposing the EQPO objective by group, the per-group gradient is scaled by $1/T_g$ where $T_g = \sqrt{N_g}\cdot\bar{r}_g$. Since the per-token surrogate loss is linear in the advantage, the scaling factors out, and the share of the optimization budget allocated to the minority group rises from $N_a/(N_a+N_b)$ under GRPO to $\sqrt{N_a}/(\sqrt{N_a}+\sqrt{N_b})$ under EQPO. Under a 100:1 imbalance, for example, this provides a $\sim 9.2\times$ improvement in the optimization budget for minority groups.

\textbf{Reward Design.} EQPO works with arbitrary reward designs. In the experiment of this work, we employ a standard accuracy reward where the model gets a reward of 1 if the final answer is correct, and a reward of 0 otherwise. Experiments with more complex rewards are included in App. \ref{app:judge_reward}.

\vspace{-0.3em}
\section{Experimental Setup}
\vspace{-0.3em}
\label{sec:experiment_setup}

\begin{table*}[t]
\centering
\small
\setlength{\tabcolsep}{3pt}
\caption{\textbf{List of Experimental Datasets.} We use 7 datasets across 5 clinical modalities. The performance metrics are an unweighted average of datasets across classes, as described in Sec. \ref{sec:experiment_setup}.
}
\label{tab:our_dataset}
\vspace{-0.3em}
\begin{adjustbox}{max width=\textwidth, max totalheight=\textheight, keepaspectratio}
{
\setlength{\tabcolsep}{3pt}
\begin{tabular}{@{}l l l l p{8.5cm} p{1.5cm}@{}}
\toprule
\textbf{Dataset} & \textbf{\# samples} & \textbf{Clinical domain} & \textbf{Modality} & \textbf{Labels} & \textbf{Demo.} \\
\midrule
CheXpert & 212K & Radiology & Chest X-ray & Atelectasis, Cardiomegaly, Consolidation, Edema, Enlarged Cardiomediastinum, Fracture, Lung
Lesion, Lung Opacity, Pleural Effusion, Pneumonia, Pneumothorax, Pleural Other, Support
Devices, No Finding & Age, Sex \\
Hemorrhage & 2.5K & Radiology & CT & No Hemorrhage, Has Hemorrhage & Age, Sex \\
VinDr-Mammo & 20K & Radiology, Oncology & Mammography & BI-RAD 1-5 & Age \\
ISIC-2020 & 33K & Dermatology, Oncology & Dermoscopy & Malignant, Benign & Age, Sex \\
HAM10000 & 10K & Dermatology, Oncology & Dermoscopy & Melanoma (MEL), Nevus (NV), Basal Cell Carcinoma (BCC), Actinic Keratosis/Intraepithelial
Carcinoma (AKIEC), Other (OTHER) & Age, Sex \\
PAD-UFES-20 & 2.3K & Dermatology, Oncology & Dermoscopy & Melanoma (MEL), Nevus (NV), Basal Cell Carcinoma (BCC), Actinic Keratosis/Intraepithelial
Carcinoma (AKIEC), Other (OTHER) & Age, Sex \\
COVID-BLUES & 362 & Radiology & Ultrasound & Has COVID, No COVID & Age \\
\bottomrule
\end{tabular}
}
\end{adjustbox}
\vspace{-1em}
\end{table*}
\begin{table*}[t]
\centering
\caption{\textbf{RQ1: Fairness and performance metrics comparison against RL and fairness mitigation baselines.} For fairness metrics, lower values are better and are indicated by $\downarrow$. For performance and combined metrics, higher values are better and are indicated by $\uparrow$. Bold values indicate the best result in each column for each model. \textbf{EQPO$_{ND}$} is the ablation of \textbf{EQPO} where the model does not have access to the ground truth demographic information, and the groups are inferred entirely via clustering. We release \textbf{MedGemma} trained with \textbf{EQPO} as \textbf{EquiMedGemma}. Performance is average over 4 runs. Additional Qwen-2.5-VL-7B results are reported in App. Tab.~\ref{tab:qwen_main_result}, Per dataset metrics are included in App. Tab.~\ref{tab:detailed_demographics_repp_qwen}--\ref{tab:detailed_demographics_eqpo_medgemma}. }
\vspace{-0.3em} 
\label{tab:fairness_performance}
\resizebox{\textwidth}{!}{
\small
\setlength{\tabcolsep}{4.5pt}
\newcommand{\methodcell}[1]{\makebox[4.2cm][l]{#1}}
\begin{tabular}{l|ccccccc|cc|cc}
\toprule
\multirow{2}{*}{\methodcell{\textbf{Training Method}}} & \multicolumn{7}{c|}{\textbf{Fairness Metrics}} & \multicolumn{2}{c|}{\textbf{Perf. Metrics}} & \multicolumn{2}{c}{\textbf{Combined}} \\
\cmidrule(lr){2-8} \cmidrule(lr){9-10} \cmidrule(lr){11-12}
& \textbf{PP $\downarrow$} & \textbf{EOD $\downarrow$} & \textbf{FPR$_{\text{Diff}}$ $\downarrow$} & \textbf{$\sigma_{\text{F1}}$ $\downarrow$} & \textbf{$\Delta$F1 $\downarrow$} & \textbf{$\sigma_{\text{Acc}}$ $\downarrow$} & \textbf{$\Delta$Acc $\downarrow$} & \textbf{Acc $\uparrow$} & \textbf{F1 $\uparrow$} & \textbf{Acc$_{\text{ES}}$ $\uparrow$} & \textbf{F1$_{\text{ES}}$ $\uparrow$} \\
\midrule
\multicolumn{12}{c}{\textbf{Base Model: QoQ-Med3-8B \cite{dai2025qoq3} }} \\
\midrule
\textbf{Re++~\citep{hu2025reinforce}} & 29.41 & 8.14 & 7.09 & .0443 & .0872 & 4.66 & 9.60 & 77.33 & .2724 & 73.89 & .2609 \\
\textbf{RLOO~\citep{rloo}} & 30.03 & 6.88 & 6.55 & .0360 & .0696 & 4.75 & 9.77 & 77.40 & .2637 & 73.90 & .2546 \\
\textbf{GRPO~\citep{shao2024deepseekmath}} & \textbf{16.63} & 10.38 & 8.47 & .0467 & .0918 & 4.76 & 10.05 & 76.62 & .3008 & 73.15 & .2873 \\
\textbf{GRPO+RS~\citep{puyol2021fairness}} & 32.15 & 6.77 & 6.60 & .0406 & .0874 & 5.57 & 11.42 & 77.45 & .2767 & 73.36 & .2659 \\
\textbf{GRPO+DRO~\citep{sagawa2019distributionally}} & 25.24 & 7.00 & 6.05 & .0459 & .0956 & 5.50 & 11.17 & 77.43 & .2764 & 73.40 & .2643 \\
\midrule
\textbf{EQPO$_{ND}$} & 30.59 & 5.70 & 5.78 & .0356 & .0715 & \textbf{3.91} & \textbf{7.85} & \textbf{79.20} & .2944 & \textbf{76.22} & .2843 \\
\textbf{EQPO (EquiQoQ-Med3)} & 22.66 & \textbf{5.49} & \textbf{5.24} & \textbf{.0262} & \textbf{.0526} & 4.53 & 9.28 & 78.66 & \textbf{.3020} & 75.25 & \textbf{.2943} \\
\midrule
\multicolumn{12}{c}{\textbf{Base Model: MedGemma-4B \cite{sellergren2025medgemma}}} \\
\midrule
\textbf{Re++~\citep{hu2025reinforce}} & 20.99 & 8.75 & 5.62 & .0518 & .1033 & 4.32 & 8.82 & 78.60 & .2978 & 75.35 & .2831 \\
\textbf{RLOO~\citep{rloo}} & 23.68 & 10.37 & 5.51 & .0600 & .1170 & 4.34 & 8.84 & 80.62 & .3047 & 77.27 & .2875 \\
\textbf{GRPO~\citep{shao2024deepseekmath}} & 22.42 & 6.48 & 4.82 & .0418 & .0795 & 4.17 & 8.55 & 80.02 & .3123 & 76.82 & .2998 \\
\textbf{GRPO+RS~\citep{puyol2021fairness}} & 23.76 & 6.66 & \textbf{3.48} & .0433 & .0835 & 4.05 & 8.39 & 80.76 & .2843 & 77.62 & .2725 \\
\textbf{GRPO+DRO~\citep{sagawa2019distributionally}} & 16.04 & 7.37 & 4.99 & .0447 & .0871 & 4.36 & 8.96 & 81.19 & .3271 & 77.80 & .3009 \\
\midrule
\textbf{EQPO$_{ND}$} & 25.15 & 11.56 & 5.69 & .0547 & .1067 & \textbf{3.61} & \textbf{7.21} & 79.23 & \textbf{.3513} & 76.47 & \textbf{.3331} \\
\textbf{EQPO (EquiMedGemma)} & \textbf{11.67} & \textbf{6.66} & 5.33 & \textbf{.0383} & \textbf{.0721} & 4.08 & 8.46 & \textbf{81.83} & .3218 & \textbf{78.62} & .3100 \\
\bottomrule
\end{tabular}
}
\vspace{-1.7em}
\end{table*}

\begin{figure*}[t]
    \centering
    \includegraphics[width=\linewidth]{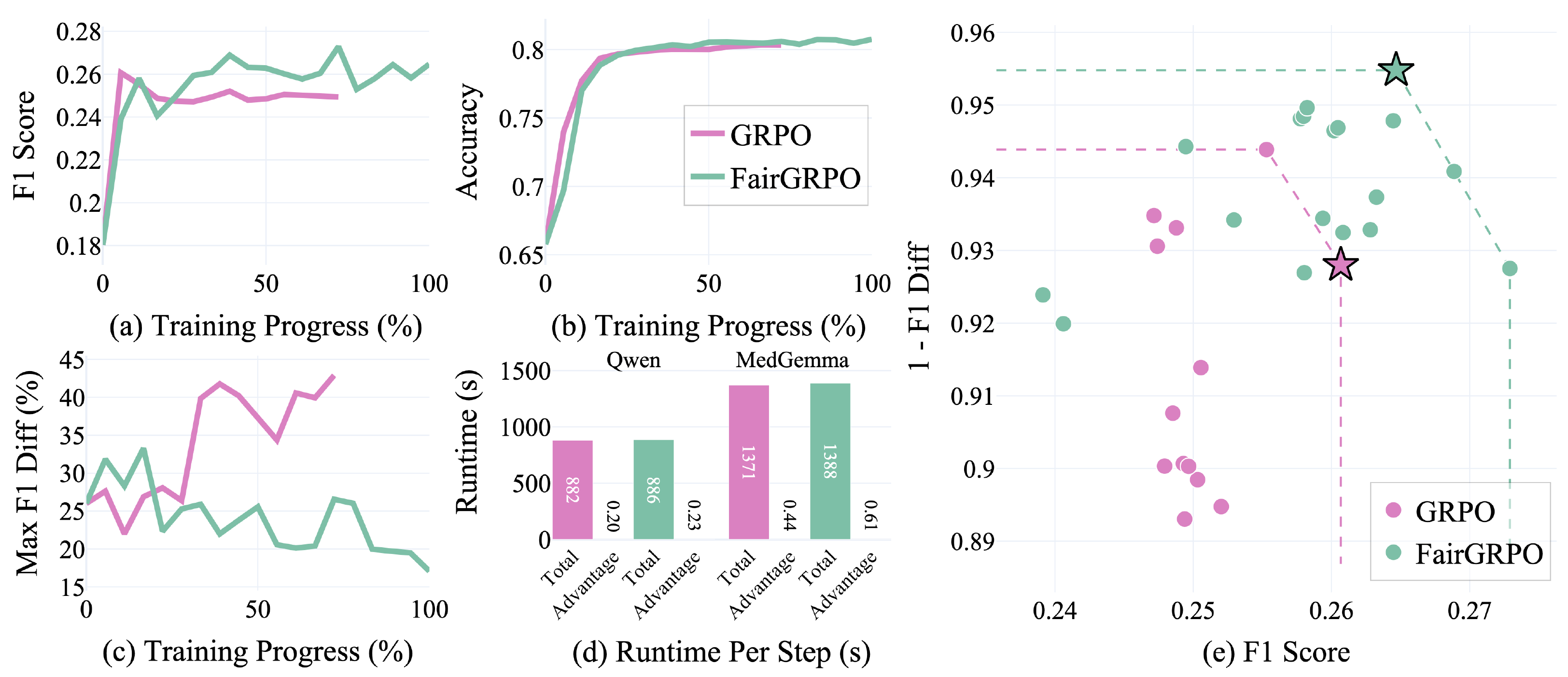}
    \caption{Training dynamics comparison between GRPO and EQPO on clinical classification tasks. \textbf{(a) F1 Score:} EQPO achieves higher F1 scores throughout training, reaching 0.265 compared to GRPO's plateau at 0.250. \textbf{(b) Accuracy:} Both methods converge to similar accuracy levels, with EQPO demonstrating slightly higher final accuracy. \textbf{(c) F1 Diff:} EQPO substantially reduces demographic performance disparities, achieving around 57\% reduction in F1 difference by explicitly optimizing for fairness during training. \textbf{(d) Per Step Runtime of the Models:} We run the model using the setup described in Sec. \ref{sec:experiment_setup}. The reward calculation for all methods is less than 0.1\% of the total runtime, showing it adds negligible overhead to the training process. \textbf{(e) Performance-Fairness Tradeoff ($\nearrow$ upper-right is better):} We compare the validation F1 score and reversed F1 difference (1-F1 Diff) of different steps throughout a single training run. Pareto frontier is plotted to illustrate the points where the model achieves the best tradeoff performance between F1 score and fairness. The starred point is the final model reported in Tab. \ref{tab:fairness_performance}. EQPO achieves superior Pareto optimality, simultaneously improving both performance and fairness compared to GRPO's best checkpoint. }
    \label{fig:training_metrics}
    \vspace{-1.3em}
\end{figure*}

Our experiments address the following three research questions:

\vspace{-0.1em}

\noindent\textbf{RQ1: How does EQPO perform compared to other 
RL methods?} We benchmark EQPO against critic-free RL baselines GRPO \citep{shao2024deepseekmath}, RLOO \cite{rloo}, and REINFORCE++ \citep{hu2025reinforce}, which represent the current state-of-the-art for critic-free RL alignment of LLMs, and against the popular bias-mitigation methods Group DRO \citep{sagawa2019distributionally} and Resampling \cite{puyol2021fairness} layered on top of GRPO.

\noindent\textbf{RQ2: How do fairness metrics evolve during training?} We track the maximum F1 score difference across demographic subgroups at 5-step intervals throughout training, and compare the trajectories against standard GRPO to test whether EQPO's adaptive weighting reshapes the optimization landscape rather than only affecting the final checkpoint.

\noindent\textbf{RQ3: How does performance vary across individual demographic groups?} We examine average F1 scores for each demographic subpopulation to test whether minority group gains come at the expense of majority group performance.

To demonstrate generalizability across architectures, we implement EQPO on three widely used VLLMs: Qwen-2.5-VL-7B \citep{bai2025qwen2}, MedGemma-4B \citep{sellergren2025medgemma}, and QoQ-Med3 \cite{dai2025qoq3}. We initialize from pretrained weights and perform unified finetuning across 7 clinical datasets in a single training run. Experiments utilize 4 NVIDIA H200 GPUs; hyperparameters are in App.~\ref{app:hyperparameters}.

\noindent\textbf{Datasets.} To ensure our methods work across different clinical datasets, we evaluate the models via 7 public datasets, including CheXpert \citep{CheXpert}, COVID-BLUES \citep{COVID_BLUES}, VinDr-Mammo \citep{VinDr-Mammo}, ISIC-2020 \citep{ISIC-2020}, HAM10000 \citep{HAM10000}, PAD-UFES-20 \citep{pacheco2020padufes20} and Hemorrhage \citep{hemorrhage}, with a total of 280.2K samples, as summarized in Tab. \ref{tab:our_dataset} and detailed in Appendix \ref{app:dataset_details}. For demographic groups, we use the patient gender as recorded in each dataset, and bin age into four 25-year brackets to maintain sufficient sample size per group: a1 (18--25), a2 (26--50), a3 (51--75), and a4 (76+).

\noindent\textbf{Evaluation Metrics.} For performance, we use hierarchical averaging of F1 scores across classes, demographic groups, and datasets. For fairness, following \citep{hort2024bias}, we report Equal Opportunity Difference (EOD), Predictive Parity (PP), and performance-variance metrics ($\sigma_{\text{F1}}$, $\Delta\text{F1}$). For the fairness-utility tradeoff, following \citep{jin2024fairmedfm}, we report Equity Scaling metrics (F1$_{\text{ES}}$, Acc$_{\text{ES}}$) that penalize models with large demographic disparities. Full definitions are in App.~\ref{app:eval_metrics}.

\noindent\textbf{Released Models.} As an artifact of this work, we publicly release two equitability-aware clinical VLLMs trained with EQPO. \textbf{EquiMedGemma-4B} is fine-tuned from MedGemma-4B \citep{sellergren2025medgemma}, and \textbf{EquiQoQ-Med3-8B} is fine-tuned from QoQ-Med3-8B \cite{dai2025qoq3}; both are produced by a unified single-pass run on the 7 clinical datasets in Tab.~\ref{tab:our_dataset}. We release the model weights, training pipeline, and evaluation scripts alongside the demographic-stratified result splits used in this paper.

\section{Results \& Discussion}

\subsection{RQ1: How does EQPO perform compared to other RL methods?}

Tab.~\ref{tab:fairness_performance} compares EQPO against RL and fairness-mitigation baselines on QoQ-Med3-8B and MedGemma-4B; Qwen-2.5-VL-7B results are in App. Tab.~\ref{tab:qwen_main_result}. On QoQ-Med3-8B, EQPO improves EOD by 47.1\%, $\sigma_{\text{F1}}$ by 43.9\%, and $\Delta\text{F1}$ by 42.7\% over vanilla GRPO, and achieves the best F1$_{\text{ES}}$ across all baselines. On MedGemma-4B, EquiMedGemma-4B reaches 27.2\% better predictive parity than Group DRO and the highest accuracy in the table (81.83\%), confirming that fairness gains do not sacrifice overall performance.

EQPO$_{ND}$ shows that our method improves fairness and performance even without demographic information during training. On QoQ-Med3-8B it achieves the best $\sigma_{\text{Acc}}$, $\Delta\text{Acc}$, accuracy, and Acc$_{\text{ES}}$ in the table; on MedGemma-4B it improves $\Delta\text{Acc}$ by 14.1\% over GRPO+RS and F1 by 12.5\% over GRPO. This suggests that reward-based clustering effectively identifies challenging samples that correlate with underrepresented demographics.

\vspace{-0.5em}
\subsection{RQ2: How does fairness evolve during training?}

\begin{figure}[t]
    \centering
    \includegraphics[width=\linewidth]{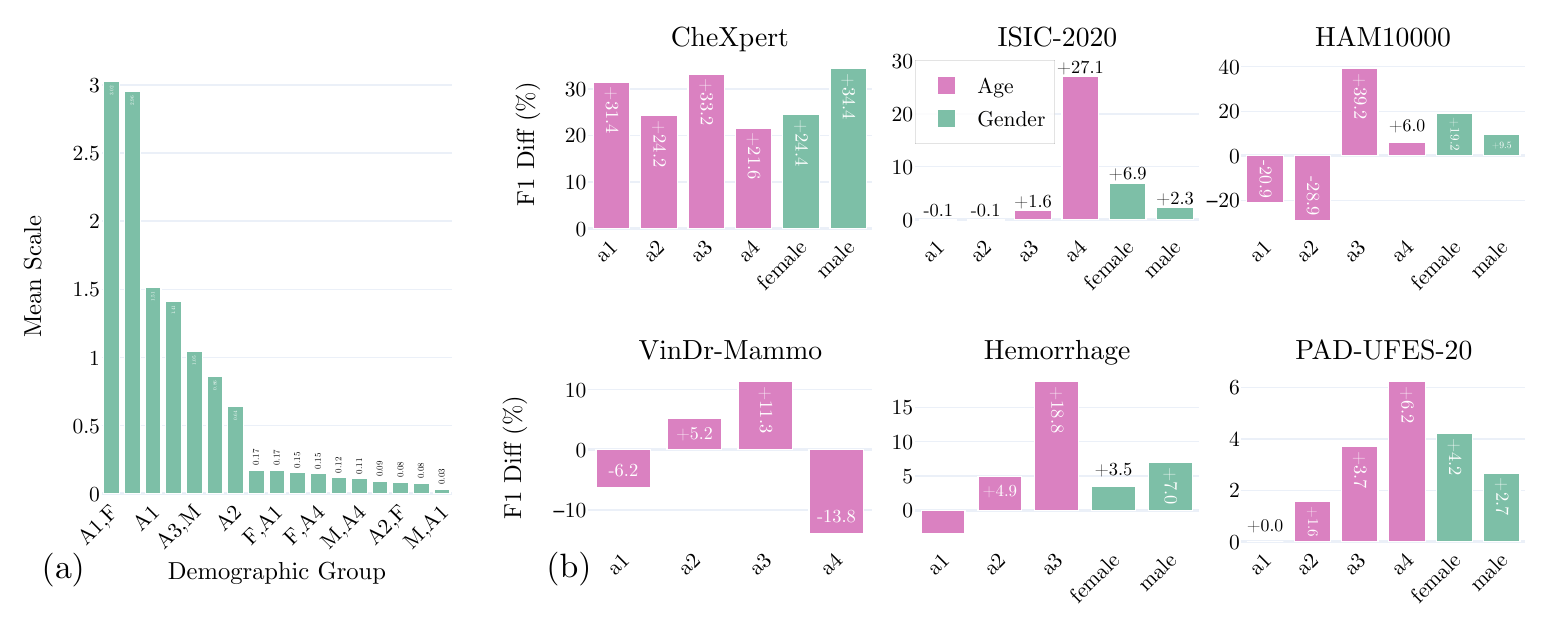}
    \vspace{-1.5em}
    \caption{\textbf{Per-group analysis of EQPO. (a) Mean EQPO$_{ND}$ scale per demographic group}, sorted by scale. Even without ground-truth demographic information, EQPO upweights minority samples (A1,F: 3.024; A4: 1.414) versus majority samples (M,A1: 0.031). \textbf{(b) F1 score differences between EQPO and GRPO} across demographic groups on MedGemma, with one panel per dataset; positive values mean EQPO performs better for the given group. EQPO improves performance on 25 of 33 demographic groups, including both majority and minority groups. Raw values are in App. Tab.~\ref{tab:group_scores_raw}.}
    \label{fig:per_group_analysis}
    \vspace{-1em}
\end{figure}

As shown in Fig.~\ref{fig:training_metrics}(c), the F1 difference for EQPO is consistently lower than that of GRPO, and the gap widens as training progresses. This divergence suggests that standard GRPO optimization increasingly favors majority groups over time, while EQPO's adaptive weighting actively counteracts this tendency. Meanwhile, Fig.~\ref{fig:training_metrics}(a,b) shows that EQPO's F1 score is higher than GRPO's while accuracy remains comparable, so the simultaneous improvement in both F1 and fairness contradicts the common assumption that fairness necessarily trades off against performance. Fig.~\ref{fig:training_metrics}(e) further shows that EQPO expands the empirical Pareto frontier relative to GRPO and provides multiple optimal checkpoints at different fairness-performance tradeoffs, all dominating GRPO's best checkpoint.

\noindent\textbf{Runtime Efficiency.} Fig.~\ref{fig:training_metrics}(d) shows EQPO and GRPO's runtime per step is close on both backbones. In particular, for all critic-free RL methods, the time for advantage calculation is less than 0.1\% of the total training time. This reveals that the extra calculation in EQPO adds \textit{negligible runtime overhead}, making it a practical drop-in replacement for standard GRPO in clinical applications.

\begin{figure}[t]
    \centering
    \includegraphics[width=\linewidth]{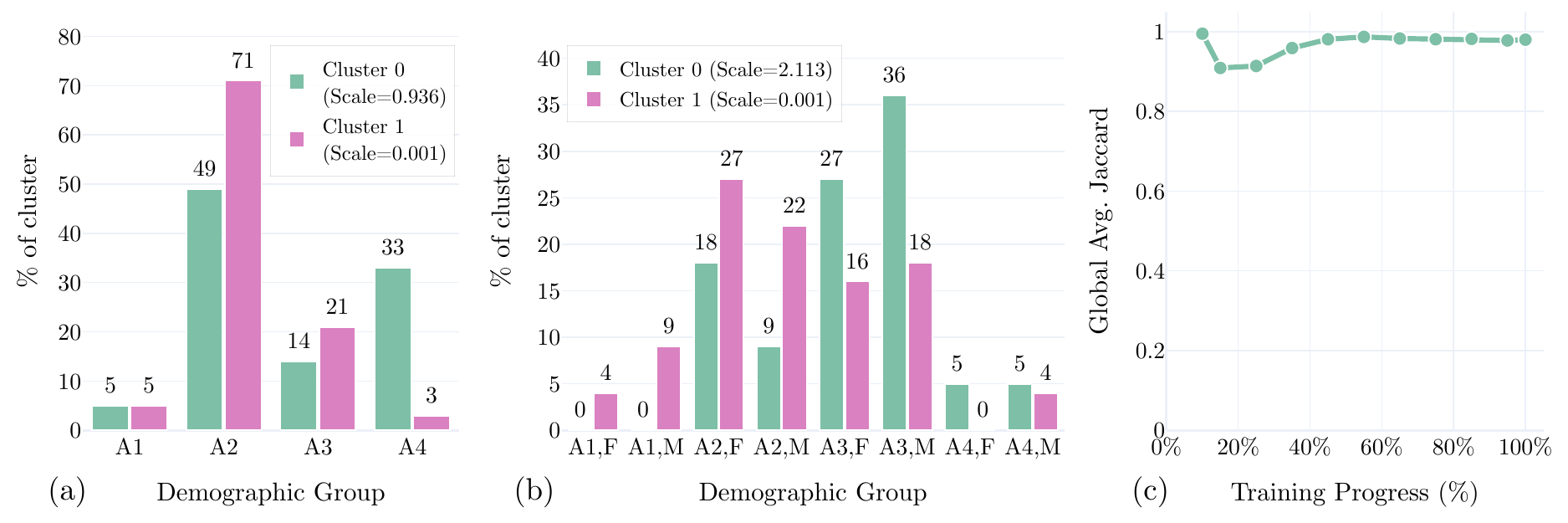}
    \caption{\textbf{(a) VinDr-Mammo} clusters split by age (A1: 18--25, A2: 26--50, A3: 51--75, A4: 75+). Cluster 0 contains 33\% A4 versus 3\% in cluster 1, with scaling favoring older groups. \textbf{(b) HAM10000} clusters split by age and gender (F = female, M = male). Cluster 0 is dominated by older patients (A3+A4 = 73\%) while cluster 1 skews younger. The scaling strongly favors older population groups. \textbf{(c) Cluster stability} measured by global average Jaccard similarity of assigned implicit groups between consecutive steps. The assignment stabilizes above $0.96$ from 35\% of training onward.}
    \label{fig:cluster_contingency}
    \vspace{-1em}
\end{figure}

\subsection{RQ3: How does performance vary across individual demographic groups?}

As shown in Fig.~\ref{fig:per_group_analysis}(b) and App. Tab.~\ref{tab:group_scores_raw}, EQPO improves performance for both underrepresented and non-underrepresented groups. In CheXpert, EQPO's F1 score is 24.4\% higher for females and 34.4\% higher for males compared to GRPO, with similarly strong gains on the youngest age groups (31.45\% on a1 and 24.32\% on a2). In PAD-UFES-20, performance on 75+ patients improves by 6.33\%; in Hemorrhage, performance on the 51--75 group improves by 18.70\%. The pattern is consistent across datasets: EQPO delivers gains on elderly subgroups while showing minimal, if any, performance degradation on younger ones, demonstrating that the fairness improvements were not achieved at the expense of majority-group performance.

\noindent\textbf{Ablations.} We further explored the effect of the cluster-count limit, the form of the temperature scaling, and the choice of clustering procedure on EQPO's behavior; full results are in App.~\ref{app:ablations}. In summary, we found that EQPO's default settings ($k=10$, sublinear $1/\sqrt{N_g}$ scaling, and reward-based clustering) deliver the best fairness-utility trade-off, while linear scaling triggers training instability and random clustering underperforms reward-based clustering on every fairness metric.

\vspace{-0.5em}
\section{Analysis}
\vspace{-0.5em}
\label{sec:analysis}
\label{sec:cluster_stability}

\begin{figure}[t]
    \centering
    \includegraphics[width=\linewidth]{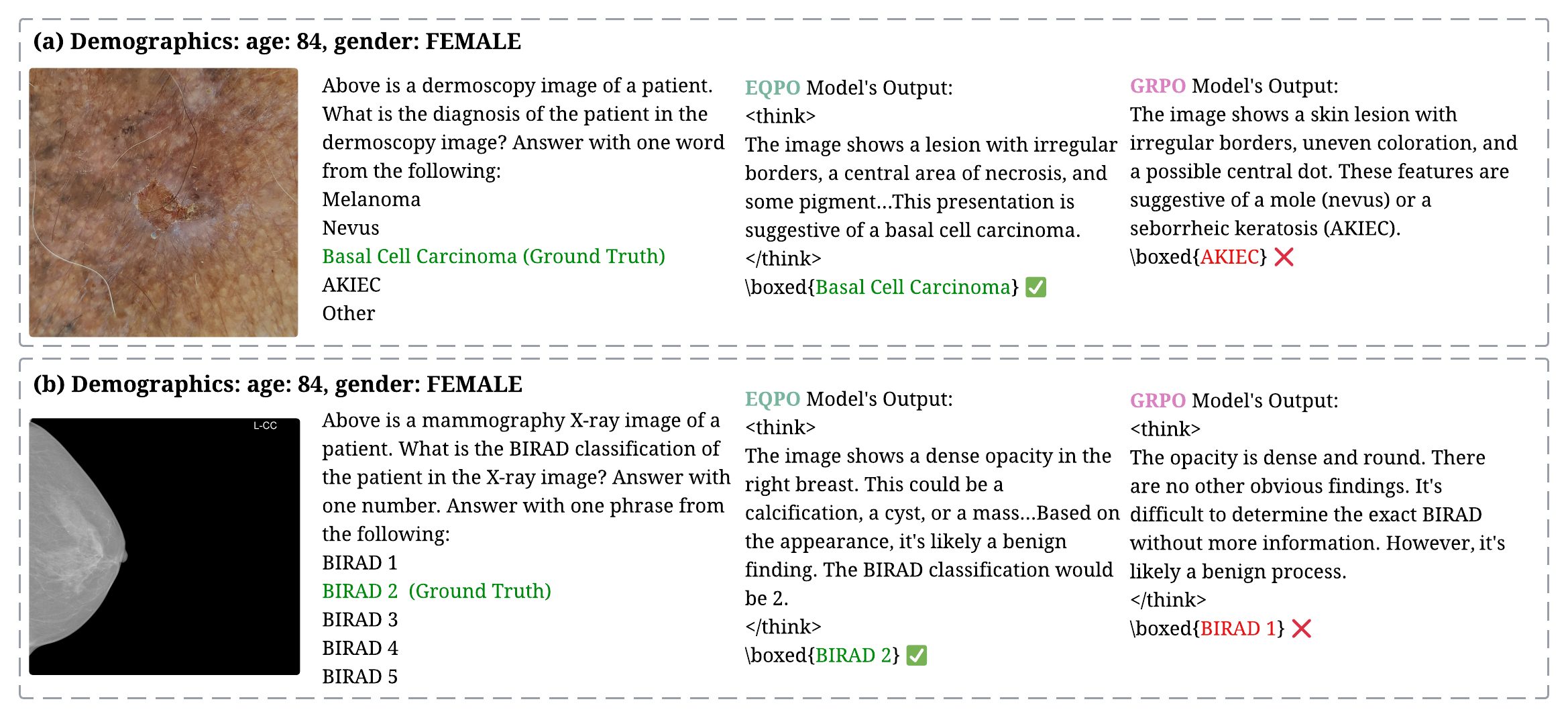}
    \vspace{-1.5em}
    \caption{\textbf{Qualitative Examples of Model's Reasoning Traces.} We see the greatest performance boosts from underrepresented groups, including samples from older population and females. In particular, models trained with EQPO exhibit an accuracy improvement of 73.08\% on 75+ populations in PAD-UFES-20 dataset, and an improvement of 36.53\% on samples aged 51-75 in VinDr-Mammo.}
    \label{fig:qualitative}
    \vspace{-1em}
\end{figure}

We complement the baseline-comparison results in §\ref{sec:experiment_setup} with analyses of the implicit groups discovered by reward-based clustering and a qualitative look at reasoning behavior.

\textbf{Quality of Implicit Group Assignment.}
With all demographic attributes masked at training time, the discovered groups achieve an average Adjusted Rand Index of 0.109 and Normalized Mutual Information of 0.355, indicating \textit{positive correlation with the ground-truth demographic categories}. Fig.~\ref{fig:per_group_analysis}(a) shows that EQPO upscales young females (A1,F: 3.024) and older age groups (A4: 1.414) most aggressively while majority groups receive scales near zero, naturally prioritizing underrepresented groups even when explicit demographic information is unavailable.

\textbf{Cluster Composition and Stability.}
Although demographic labels are masked during the EQPO$_{ND}$ run, the resulting clusters separate cleanly along the demographic axis (Fig.~\ref{fig:cluster_contingency}(a,b)). In both VinDr-Mammo and HAM10000, the elderly-enriched clusters receive mean EQPO$_{ND}$ scales of 2.113 and 0.936 versus 0.001 for their majority counterparts (App. Tab.~\ref{tab:cluster_scale}), recovering demographically meaningful structure and amplifying the underrepresented portion without supervision. We further measure Jaccard co-assignment similarity between consecutive training steps. As shown in Fig.~\ref{fig:cluster_contingency}(c), after a brief early dip the assignment stabilizes above 0.96 from 35\% of training onward, with six of seven domains converging to Jaccard 1.000 at the end of training. 

\textbf{Qualitative Analysis.}
EQPO produces stronger diagnostic reasoning than GRPO on underrepresented populations, where GRPO is more prone to hallucinated or unevidenced features. In Fig.~\ref{fig:qualitative}(a), on an 84-year-old female's dermoscopy image, EQPO correctly diagnoses Basal Cell Carcinoma from irregular borders and central necrosis, whereas GRPO hallucinates a non-existent \textit{central dot} and misdiagnoses AKIEC. In Fig.~\ref{fig:qualitative}(b), on an elderly female's mammography, EquiMedGemma correctly contextualizes a dense opacity as BI-RADS 2, while the GRPO-trained model underestimates severity and misclassifies it as BI-RADS 1. Beyond the quantitative metrics, EQPO's equitability-aware training improves clinical reasoning quality on historically underserved demographic groups.
\vspace{-1em}
\section{Conclusion}
\vspace{-0.5em}

In this work, we introduced EQPO, a novel reinforcement learning approach that addresses demographic disparities in clinical vision-language models. EQPO integrates adaptive importance weighting directly into the advantage estimation process, ensuring minority and underrepresented groups receive equitable learning signals during training. EQPO's clustering further enables equitability optimization even when demographic labels are unavailable, discovering latent groups that correlate with ground-truth demographics and remain stable throughout training. Evaluation across 7 clinical datasets spanning 5 modalities shows that EQPO reduces F1 standard deviation by 43.9\% and the maximum cross-group F1 gap by 42.7\% on QoQ-Med3-8B over vanilla GRPO, and narrows predictive parity disparities by 27.2\% on MedGemma-4B over bias-mitigated baselines while raising F1 by 12.5\%. Through the release of EquiMedGemma-4B and EquiQoQ-Med3-8B, we provide the first publicly available clinical VLLMs explicitly optimized for demographic equitability through RL. By establishing equitability as a fundamental optimization objective, we hope this work contributes toward AI-assisted diagnostic systems that serve all patient populations equitably.

\bibliography{example_paper}
\bibliographystyle{plain}


\appendix

\section{Hyperparameters \& Setups}
\label{app:hyperparameters}

In this section, we describe our setup and hyperparameters during the training of the model. All models are trained with 4 NVIDIA H200 GPUs.

\begin{table}[ht]
\centering
\caption{Hyperparameters for All Trainings}
\label{tab:hyperparameters}
\begin{tabular}{ll}
\toprule
\textbf{Parameter} & \textbf{Value} \\
\midrule
\multicolumn{2}{l}{\textit{Data Configuration}} \\
Train batch size & 512 \\
Validation batch size & 512 \\
Max prompt length & 4096 \\
Max response length & 4096 \\
\midrule
\multicolumn{2}{l}{\textit{Model Configuration}} \\
Base model & MedGemma-4B-IT/Qwen2.5-VL-7B-Instruct \\
Tensor model parallel size & 2 \\
\midrule
\multicolumn{2}{l}{\textit{Optimization}} \\
Learning rate & $5 \times 10^{-7}$ \\
PPO mini-batch size & 128 \\
PPO micro-batch size per GPU & 4 \\
KL & Disabled \\
\midrule
\multicolumn{2}{l}{\textit{Rollout Configuration}} \\
Number of rollouts ($n$) & 10 \\
GPU memory utilization & 0.6 \\
Rollout engine & VLLM \\
\midrule
\multicolumn{2}{l}{\textit{Training Settings}} \\
Total epochs & 15 \\
Validation frequency & 5 epochs \\
Model save frequency & 20 steps \\
Number of GPUs per node & 4 \\
Number of nodes & 1 \\
Critic warmup steps & 0 \\
\bottomrule
\end{tabular}
\end{table}

All experiments were conducted using the VERL (Volcano Engine Reinforcement Learning for LLMs) framework. The model was initialized from the pretrained MedGemma-4B-IT checkpoint and fine-tuned. We employed vLLM for efficient rollout generation with a GPU memory cache of 60\% to balance between batch size and memory constraints. The relatively low learning rate of $5 \times 10^{-7}$ was chosen to ensure stable convergence given the complexity of the multi-task medical reasoning objective.

\subsection{Evaluation Metrics}
\label{app:eval_metrics}

To comprehensively evaluate both performance and fairness across heterogeneous clinical subpopulations, we employ a hierarchical evaluation framework that prevents any single dataset or demographic subgroup from dominating the assessment.

\textbf{Notation.} Let $\mathcal{C}_k$ denote the set of classes for dataset $k$, and $\mathcal{G}$ denote the set of demographic groups. For each class $c \in \mathcal{C}_k$ and group $g \in \mathcal{G}$, we define: $TP_{c,g}$ (true positives), $FP_{c,g}$ (false positives), $TN_{c,g}$ (true negatives), and $FN_{c,g}$ (false negatives). Let $n_{c,g}$ denote the number of samples for class $c$ in group $g$.

\textbf{Performance Metrics.} We extract diagnoses from the model's free-text reasoning traces and evaluate each class as a binary classification problem. For class $c$ and group $g$:

\begin{align}
\text{Acc}_{c,g} &= \frac{TP_{c,g} + TN_{c,g}}{n_{c,g}}, \quad &\text{Precision}_{c,g} &= \frac{TP_{c,g}}{TP_{c,g} + FP_{c,g}} \\
\text{Recall}_{c,g} &= \frac{TP_{c,g}}{TP_{c,g} + FN_{c,g}}, \quad &\text{F1}_{c,g} &= 2 \cdot \frac{\text{Precision}_{c,g} \cdot \text{Recall}_{c,g}}{\text{Precision}_{c,g} + \text{Recall}_{c,g}}
\end{align}

To ensure balanced representation across classes and datasets, we employ two-level averaging. For dataset $k$:
\begin{align}
\text{F1}_k = \frac{1}{|\mathcal{C}_k|} \sum_{c \in \mathcal{C}_k} \text{F1}_c, \quad \text{where} \quad \text{F1}_c = \frac{1}{|\mathcal{G}|} \sum_{g \in \mathcal{G}} \text{F1}_{c,g}
\end{align}

The overall performance is then averaged across all $K$ datasets:
\begin{align}
\text{F1}_{\text{overall}} = \frac{1}{K} \sum_{k=1}^{K} \text{F1}_k
\end{align}

This hierarchical averaging ensures that no single class or dataset dominates the final metrics, allowing the final metrics to be a balanced assessment across all 5 clinical domains.

\textbf{Fairness Metrics.} Following the popular approaches outlined in \citep{hort2024bias}, we evaluate fairness through multiple complementary perspectives, each capturing different aspects of equitable model behavior across demographic groups. For each metric, we first compute dataset-level performance for each group, then assess disparities across groups.

\textit{Equal Opportunity Difference (EOD):} We measure the disparity in true positive rates across groups to ensure equal diagnostic sensitivity:
\begin{align}
\text{EOD} = \max_{g \in \mathcal{G}} \text{TPR}_{g} - \min_{g \in \mathcal{G}} \text{TPR}_{g}, \quad \text{where} \quad \text{TPR}_{g} = \frac{1}{K} \sum_{k=1}^{K} \frac{1}{|\mathcal{C}_k|} \sum_{c \in \mathcal{C}_k} \text{TPR}_{c,g}
\end{align}

and $\text{TPR}_{c,g} = \frac{TP_{c,g}}{TP_{c,g} + FN_{c,g}}$. A lower EOD indicates more equitable identification of positive cases, which is crucial for preventing delayed diagnoses in underserved populations.

\textit{Predictive Parity:} We assess the reliability of positive predictions across groups through false discovery rate gaps:
\begin{align}
\text{PP} = \max_{g \in \mathcal{G}} \text{FDR}_{g} - \min_{g \in \mathcal{G}} \text{FDR}_{g}, \quad \text{where} \quad \text{FDR}_{g} = \frac{1}{K} \sum_{k=1}^{K} \frac{1}{|\mathcal{C}_k|} \sum_{c \in \mathcal{C}_k} \text{FDR}_{c,g}
\end{align}

and $\text{FDR}_{c,g} = \frac{FP_{c,g}}{FP_{c,g} + TP_{c,g}}$. Lower predictive parity gaps ensure that positive predictions maintain consistent reliability across all demographic groups, fostering trust in AI-assisted diagnosis.

\textit{False Positive Rate Difference:} We measure disparities in false positive rates to ensure equitable specificity across groups:
\begin{align}
\text{FPR}_{\text{Diff}} = \max_{g \in \mathcal{G}} \text{FPR}_{g} - \min_{g \in \mathcal{G}} \text{FPR}_{g}
\end{align}
where $\text{FPR}_{g}$ follows the same hierarchical averaging structure as other group-level metrics. Lower FPR differences prevent differential overdiagnosis across demographic groups.

\textit{Performance Disparities:} We directly measure accuracy and F1 score gaps to capture overall performance equity:
\begin{align}
\Delta\text{Acc} &= \max_{g \in \mathcal{G}} \text{Acc}_{g} - \min_{g \in \mathcal{G}} \text{Acc}_{g}, \quad &\Delta\text{F1} &= \max_{g \in \mathcal{G}} \text{F1}_{g} - \min_{g \in \mathcal{G}} \text{F1}_{g}
\end{align}

where $\text{Acc}_{g}$ and $\text{F1}_{g}$ follow the same hierarchical averaging as TPR$_g$. Additionally, we compute the standard deviation of performance across groups to capture variability:
\begin{align}
\sigma_{\text{Acc}} &= \sqrt{\frac{1}{|\mathcal{G}|} \sum_{g \in \mathcal{G}} (\text{Acc}_{g} - \overline{\text{Acc}})^2}, \quad &\sigma_{\text{F1}} &= \sqrt{\frac{1}{|\mathcal{G}|} \sum_{g \in \mathcal{G}} (\text{F1}_{g} - \overline{\text{F1}})^2}
\end{align}

where $\overline{\text{Acc}}$ and $\overline{\text{F1}}$ denote the mean values across all groups.

\textbf{Fairness-Utility Tradeoff.} To balance fairness and utility, we adopt Equity Scaling metrics following \citep{jin2024fairmedfm}. These metrics combine performance with fairness considerations by penalizing models that achieve high average performance at the cost of large disparities across groups:
\begin{align}
\text{Acc}_{\text{ES}} &= \frac{\overline{\text{Acc}}}{1 + \sigma_{\text{Acc}}}, \quad &\text{F1}_{\text{ES}} &= \frac{\overline{\text{F1}}}{1 + \sigma_{\text{F1}}}
\end{align}

These equity-scaled metrics reward models that achieve both high performance and low variance across demographic groups, providing a single scalar that captures the fairness-utility tradeoff. Higher values indicate better balance between overall performance and equitable distribution across all populations.

\section{Dataset Details}
\label{app:dataset_details}

In this section, we provide a detailed description of datasets used in the experiments. 

\noindent\textbf{CheXpert}~\citep{CheXpert} is a public chest radiology dataset collected at Stanford Hospital, which contains 224{,}316 chest radiographs of 65{,}240 patients. Each record has an uncertain label of 14 diagnostic observations, including Atelectasis, Cardiomegaly, Consolidation, Edema, Enlarged Cardiomediastinum, Fracture, Lung Lesion, Lung Opacity, Pleural Effusion, Pneumonia, Pneumothorax, Pleural Other, Support Device and No Finding. We use a training set of 212,243 records, a test set of 225 records, and a total size of 212,498 records.

\noindent\textbf{COVID-BLUES}~\citep{COVID_BLUES} consists of bluepoint-specific lung ultrasound videos collected at the Maastricht University Medical Center in the Netherlands using the BLUE protocol. Each of the 63 patients has six recordings. Our evaluation focuses on two labels: the diagnostic label (``Has COVID'', ``No COVID''), and the patient age label. We use a training set of 266 records, a test set of 96 records, and a total size of 362 records.

\noindent\textbf{VinDr-Mammo}~\citep{VinDr-Mammo} contains mammography collected from Hospital 108 and Hanoi Medical University Hospital in Vietnam. The dataset includes local labels for bounding boxes; however, we evaluate our models based on the 5 global labels for BI-RADS 1-5. We use a training set of 16{,}000 records, a test set of 4{,}000 records, and a total size of 20{,}000 records.

\noindent\textbf{ISIC-2020}~\citep{ISIC-2020} comprises dermoscopy of skin lesions from over 2{,}000 patients, generated by the International Skin Imaging Collaboration (ISIC). We evaluate the models on the binary classification (``Malignant'' or ``Benign'') for each image, where all malignant diagnoses are histopathology\textendash confirmed, while benign diagnoses are confirmed by expert agreement, longitudinal follow\textendash up, or histopathology.We use a training set of 26{,}501 records, a test set of 6{,}625 records, and a total size of 33{,}126 records.

\noindent\textbf{HAM10000}~\citep{2018ham10000} is a dermoscopic image dataset released for the ISIC 2018 classification challenge, drawn from the ISIC archive. Our evaluation uses the diagnostic categories: Melanoma (MEL), Nevus (NV), Basal Cell Carcinoma (BCC), Actinic Keratosis/Intraepithelial Carcinoma (AKIEC), Other (OTHER).We use a training set of 8,012 records, a test set of 2,003 records, and a total size of 10,015 records.

\noindent\textbf{PAD-UFES-20}~\citep{pacheco2020padufes20} comprises dermoscopy images of skin lesions with patient metadata collected at the Federal University of Espírito Santo by iPhone, which includes 1{,}641 skin lesions from 1{,}373 patients. We evaluate the models on the five skin diagnostics, three of which are skin disease and three of which are skin cancers: Melanoma (MEL), Nevus (NV), Basal Cell Carcinoma (BCC), Actinic Keratosis/Intraepithelial Carcinoma (AKIEC), Other (OTHER). All of the skin cancers are biopsy-proven, and more than half of the skin diseases are biopsy-proven as well. We use a training set of 1{,}839 records, a test set of 459 records, and a total size of 2{,}298 records.

\noindent\textbf{Hemorrhage}~\citep{hemorrhage} consists of intracranial hemorrhage CT images for 82 patients at Al Hilla Teaching Hospital, Iraq, each with brain and bone window images and approximately 30 image slices in total. We evaluate the models as binary diagnoses: ``No Hemorrhage'' and ``Has Hemorrhage''. We use a training set of 1{,}986 records, a test set of 515 patient records, and a total size of 2{,}501 records.

\section{Convergence Analysis}
\label{app:theory}

\textit{Setup.} Consider groups $g \in \{g_1, \ldots, g_G\}$ where $g_1$ is minority and $g_2$ is majority, with sample counts $N_{g_1} \ll N_{g_2}$, total $N = \sum_g N_g$, and mean reward $\bar{r}_{g,t}$ at iteration $t$. We use accuracy reward $r \in \{0,1\}$, so $\bar{r}_{g,t} = \text{Acc}_{g,t}$. For simplicity, assume all groups start with roughly equal performance: $\bar{r}_{g,0} \approx \bar{r}_0$ for all $g$. The fairness metric we aim to minimize is $\sigma_{\text{Acc}} = \sqrt{\frac{1}{|\mathcal{G}|}\sum_g(\text{Acc}_g - \overline{\text{Acc}})^2}$.

\subsection{Per-step gradient allocation ($\beta = 0$)}

The EQPO objective is $J(\theta) = \mathbb{E}_{q,o}[\sum_k L_k(\theta) - \beta D_{\text{KL}}(\pi_\theta \| \pi_{\text{ref}})]$, where $L_k(\theta) = \min(\varphi_k(\theta)\hat{A},\ \text{clip}(\varphi_k(\theta), 1\pm\varepsilon)\hat{A})$. Here $k$ indexes over rollouts, $\varphi_k(\theta) = \pi_\theta(o_k \mid q) / \pi_{\theta_{\text{old}}}(o_k \mid q)$ is the standard importance sampling ratio inherited from GRPO, and $\hat{A}$ is the advantage. Since the importance ratio $\varphi_k$ is shared between GRPO and EQPO and is unrelated to our fairness temperature scaling, it plays no role in the analysis below. Additionally, as shown in Appendix~\ref{app:hyperparameters}, we disable the KL divergence penalty in all experiments, setting $\beta = 0$. We therefore focus on the surrogate loss without the KL term.

Decomposing by group gives
\begin{equation}
J(\theta) = \sum_g \sum_{q \in g}\sum_k L_k(\theta; \hat{A}_{q,k}).
\end{equation}
In EQPO, $\hat{A}_{q,k}^{\text{EQPO}} = \hat{A}_{q,k}^{\text{GRPO}} / (T_{g(q)} \cdot \sigma_{\text{batch}})$, where $T_g = \sqrt{N_g}\cdot\bar{r}_{g,t}$ is the temperature and $\sigma_{\text{batch}}$ is the batch normalization constant. Since $L_k$ is linear in $\hat{A}$, the scaling extracts as
\begin{equation}
L_k(\theta; \hat{A}_{q,k}^{\text{EQPO}}) = \frac{1}{T_{g(q)}\cdot\sigma_{\text{batch}}} L_k(\theta; \hat{A}_{q,k}^{\text{GRPO}}).
\end{equation}

At $t=0$, since $\bar{r}_{g,0} \approx \bar{r}_0$ for all $g$, the temperature reduces to $T_g \approx \sqrt{N_g}\cdot\bar{r}_0$, so $T_{g_2} = \sqrt{N_{g_2}}\cdot\bar{r}_0 > \sqrt{N_{g_1}}\cdot\bar{r}_0 = T_{g_1}$ and $1/T_{g_1} > 1/T_{g_2}$. Taking the gradient, noting that $1/T_g$ is constant with respect to $\theta$ within an iteration, gives
\begin{equation}
\nabla_\theta J_g^{\text{EQPO}} = \frac{1}{T_g\cdot\sigma_{\text{batch}}}\nabla_\theta\sum_{q\in g}\sum_k L_k(\theta; \hat{A}_{q,k}^{\text{GRPO}}).
\end{equation}

In GRPO, no such scaling exists, so every group's $L_k$ contributes equally. The gradient contribution of group $g$ is proportional to $N_g$, meaning majority groups dominate, i.e., $\|\nabla_\theta J_{g_2}^{\text{GRPO}}\| \gg \|\nabla_\theta J_{g_1}^{\text{GRPO}}\|$. A larger gradient produces a larger per-step policy change on that group's data, yielding larger $\Delta\text{Acc}_g$. Starting from equal performance, GRPO causes majority groups to improve rapidly while minority groups see minimal improvement, so $\sigma_{\text{Acc}}^{\text{GRPO}}$ increases with each iteration.

In EQPO, the $1/T_g$ scaling counteracts the frequency imbalance. Minority group $g_1$, having small $N_g$ and thus small $T_g$, receives amplified gradients, while majority group $g_2$, having large $N_g$ and thus large $T_g$, is attenuated. All groups improve at a more equal rate, so starting from equal performance, the gap between groups remains compressed,
\begin{equation}
\sigma_{\text{Acc}}^{\text{EQPO}} = \sqrt{\frac{1}{|\mathcal{G}|}\sum_g(\text{Acc}_{g,t+1}^{\text{EQPO}} - \overline{\text{Acc}}_{t+1})^2} < \sigma_{\text{Acc}}^{\text{GRPO}}. \quad \square
\end{equation}

Therefore, EQPO strictly improves fairness as measured by $\sigma_{\text{Acc}}$ under the assumption of equal initial group performance and linear advantage scaling.

\subsection{Convergence under finite KL budget}

The $\beta = 0$ analysis above matches our experimental setup but does not address whether the front-loaded minority improvement persists once the KL term is reintroduced. Here we show that under the full objective with $\beta > 0$, the per-group allocation derived above is locked in at convergence.

GRPO normalizes advantages to zero mean and unit variance \emph{within each prompt's response group}: $\hat{A}_{q,k}^{\text{GRPO}} = (r_{q,k} - \hat{\mu}_q)/(\hat{\sigma}_q + \varepsilon)$. This guarantees $\mathbb{E}_q[\hat{A}_q] = 0$ and $\text{Var}_q[\hat{A}_q] = 1$ regardless of current accuracy, so the magnitude of the reward-side gradient is approximately constant throughout training. Meanwhile, as $\pi_\theta$ diverges from $\pi_{\text{ref}}$, $D_{\text{KL}}(\pi_\theta \| \pi_{\text{ref}})$ grows, producing an increasing restoring force. The two forces balance at a finite convergence point $T^*$ at which the cumulative KL budget $\sum_t \beta\, D_{\text{KL}}^{(t)}$ saturates the implicit penalty.

Consider a more general reward $r \in [0, R_{\max}]$ (not necessarily binary) and similar per-group difficulty $\bar{r}_g \approx \bar{r}$. Under the mild assumption that cross-group gradient alignment does not systematically favor either group, per-group cumulative improvement up to $T^*$ satisfies $\Delta\text{Acc}_g \approx C \cdot w_g \cdot N_g \cdot |\hat{A}|$, where $C$ is shared across methods (batch renormalization keeps the total gradient magnitude comparable) and $|\hat{A}|$ is the approximately constant advantage magnitude. Substituting weights $w_g^{\text{GRPO}} = 1$ and $w_g^{\text{EQPO}} = 1/T_g \approx 1/(\sqrt{N_g}\bar{r})$, the share of total improvement allocated to the minority group $g_1$ is
\begin{equation}
\rho_{g_1}^{\text{GRPO}} = \frac{N_{g_1}}{N_{g_1} + N_{g_2}}, \qquad \rho_{g_1}^{\text{EQPO}} = \frac{\sqrt{N_{g_1}}}{\sqrt{N_{g_1}} + \sqrt{N_{g_2}}}.
\label{eq:minority-share}
\end{equation}
Taking the ratio under a 100:1 imbalance ($N_{g_2}/N_{g_1} = 100$) gives
\begin{equation}
\frac{\rho_{g_1}^{\text{EQPO}}}{\rho_{g_1}^{\text{GRPO}}} = \frac{\sqrt{N_{g_1}} + N_{g_2}/\sqrt{N_{g_1}}}{\sqrt{N_{g_1}} + \sqrt{N_{g_2}}} \cdot \frac{N_{g_1}}{N_{g_1} + N_{g_2}} \cdot \frac{1}{N_{g_1}}\cdot(N_{g_1}+N_{g_2})  \;\approx\; \frac{1 + 100}{1 + 10} \;\approx\; 9.18,
\end{equation}
i.e., the minority group receives approximately $9.2\times$ more of the optimization budget under EQPO than under GRPO. Because $T^*$ is finite, this allocation cannot be eroded by subsequent updates: any later majority-favoring move would require expanding the KL budget beyond its bound. The front-loaded compression of $\sigma_{\text{Acc}}$ derived in the $\beta = 0$ case therefore persists at convergence.

Furthermore, EQPO's dynamic scaling brings two benefits.

\textbf{D1. Performance-based dynamic adjustment.} Define the effective weight $w_{g,t} = \sqrt{N_g}/(N\cdot\bar{r}_{g,t})$. If a minority group improves by $\delta_1 > 0$, the weight ratio $R_t = w_{g_1,t}/w_{g_2,t}$ satisfies $R_{t+1} < R_t$. If it stagnates while the majority improves, $R_{t+1} > R_t$. EQPO always shifts focus toward the group with lower relative improvement.

\textbf{D2. Distribution-based dynamic adjustment.} Since $T_{g,t} = \sqrt{N_{g,t}}\cdot\bar{r}_{g,t}$ is recomputed every iteration, if a group loses samples at iteration $t'$ such that $N_{g,t'} < N_{g,t'-1}$, it is immediately upweighted with $w_{g,t'} > w_{g,t'-1}$. Group DRO and resampling cannot jointly adapt to changes in both group composition and performance within the RL loop.

\section{Pseudocode for EQPO}
\label{app:pseudocode}

Algorithm~\ref{alg:grpo_adv} computes the standard GRPO advantage and Algorithm~\ref{alg:eqpo_adv} computes the EQPO advantage. EQPO differs from GRPO only in the per-group temperature scaling step and the batch renormalization at the end; the rest of the policy gradient pipeline (importance ratio, clipping, KL regularization, optimizer step) is unchanged.

\begin{algorithm}[h]
\caption{GRPO advantage computation for prompt $q$.}
\label{alg:grpo_adv}
\begin{algorithmic}[1]
\Require Prompt $q$, rollout group $G_q$ with rewards $\{r_{q,k}\}_{k=1}^{|G_q|}$
\State $\hat{\mu}_q \gets \tfrac{1}{|G_q|}\sum_k r_{q,k}$
\State $\hat{\sigma}_q \gets \mathrm{stddev}(\{r_{q,k}\})$
\For{$k = 1, \ldots, |G_q|$}
  \State $\hat{A}_{q,k}^{\text{GRPO}} \gets (r_{q,k} - \hat{\mu}_q) / (\hat{\sigma}_q + \varepsilon)$
\EndFor
\State \Return $\{\hat{A}_{q,k}^{\text{GRPO}}\}$
\end{algorithmic}
\end{algorithm}

\begin{algorithm}[h]
\caption{EQPO advantage computation for batch $\mathcal{B}$ at iteration $t$.}
\label{alg:eqpo_adv}
\begin{algorithmic}[1]
\Require Batch $\mathcal{B}$ of prompts. Each prompt $q$ has a domain $g(q)$ and a group $\gamma(q)$ that is either an explicit demographic label or, when missing, the K-means cluster assignment in domain $g(q)$ on per-prompt reward vectors at iteration $t$. Each prompt has rollout rewards $\{r_{q,k}\}$.
\For{each domain $g$}
  \State $N_{g,t}, \bar{r}_{g,t} \gets$ count and mean reward in domain $g$
  \For{each group $\gamma$ in domain $g$}
    \State $N_{\gamma,g,t}, \bar{r}_{\gamma,g,t} \gets$ count and mean reward in group $\gamma$
    \State $T_{g,t} \gets \sqrt{N_{g,t}}\cdot\bar{r}_{g,t}$, \quad $T_{\gamma,g,t} \gets \sqrt{N_{\gamma,g,t}}\cdot\bar{r}_{\gamma,g,t}$
  \EndFor
\EndFor
\For{each prompt $q \in \mathcal{B}$}
  \State $\hat{\mu}_q, \hat{\sigma}_q \gets$ mean and stddev of $\{r_{q,k}\}$
  \For{$k = 1, \ldots, |G_q|$}
    \State $s_{q,k} \gets (r_{q,k} - \hat{\mu}_q)/(\hat{\sigma}_q + \varepsilon)$
    \State $s_{q,k}^{\text{scaled}} \gets s_{q,k} / \max\!\big(T_{g(q),t}\cdot T_{\gamma(q),g(q),t},\,\varepsilon\big)$
  \EndFor
\EndFor
\State $\sigma_{\text{batch}} \gets \mathrm{stddev}\big(\{s_{q,k}^{\text{scaled}}\}_{(q,k)\in\mathcal{B}}\big)$
\For{each $(q,k)$}
  \State $\hat{A}_{q,k}^{\text{EQPO}} \gets s_{q,k}^{\text{scaled}} / \sigma_{\text{batch}}$
\EndFor
\State \Return $\{\hat{A}_{q,k}^{\text{EQPO}}\}$
\end{algorithmic}
\end{algorithm}

\section{Ablation Studies}
\label{app:ablations}

\begin{table*}[t]
\centering
\caption{\textbf{Ablations on cluster-count limit, scaling form, and clustering type, all on QoQ-Med3-8B.} \emph{Cluster count} varies the upper limit $k$ for the elbow method; $k=10$ is the EQPO default. \emph{Scaling form} compares no scaling (equivalent to vanilla GRPO), sublinear scaling $1/(\sqrt{N_g}\bar{r}_g)$ (EQPO default), and linear scaling $1/(N_g\bar{r}_g)$. \emph{Clustering type} ablates the EQPO$_{ND}$ procedure when demographic labels are masked. Lower is better for fairness metrics ($\downarrow$); higher is better for performance and combined metrics ($\uparrow$). Bold marks the best result within each block.}
\label{tab:ablations}
\resizebox{\textwidth}{!}{
\small
\setlength{\tabcolsep}{4.5pt}
\begin{tabular}{l|ccccccc|cc|cc}
\toprule
\textbf{Setting} & \textbf{PP $\downarrow$} & \textbf{EOD $\downarrow$} & \textbf{FPR$_{\text{Diff}}$ $\downarrow$} & \textbf{$\sigma_{\text{F1}}$ $\downarrow$} & \textbf{$\Delta$F1 $\downarrow$} & \textbf{$\sigma_{\text{Acc}}$ $\downarrow$} & \textbf{$\Delta$Acc $\downarrow$} & \textbf{Acc $\uparrow$} & \textbf{F1 $\uparrow$} & \textbf{Acc$_{\text{ES}}$ $\uparrow$} & \textbf{F1$_{\text{ES}}$ $\uparrow$} \\
\midrule
\multicolumn{12}{c}{\textbf{(a) Cluster count $k$}} \\
\midrule
\textbf{$k = 1$ (no clustering)} & 27.49 & 6.56 & 5.45 & .0448 & .0924 & 5.34 & 10.82 & 77.78 & .2704 & 73.84 & .2588 \\
\textbf{$k = 10$ (EQPO default)} & \textbf{22.66} & \textbf{5.49} & 5.24 & \textbf{.0262} & \textbf{.0526} & 4.53 & 9.28 & 78.66 & .3020 & 75.25 & .2943 \\
\textbf{$k = 20$} & 23.72 & 5.73 & \textbf{4.59} & .0317 & .0662 & \textbf{4.37} & \textbf{8.67} & \textbf{78.99} & \textbf{.3066} & \textbf{75.68} & \textbf{.2972} \\
\midrule
\multicolumn{12}{c}{\textbf{(b) Scaling form}} \\
\midrule
\textbf{No scaling (vanilla GRPO)} & \textbf{16.63} & 10.38 & 8.47 & .0467 & .0918 & 4.76 & 10.05 & 76.62 & .3008 & 73.15 & .2873 \\
\textbf{Sublinear $1/\sqrt{N_g}$ (EQPO)} & 22.66 & 5.49 & \textbf{5.24} & \textbf{.0262} & \textbf{.0526} & 4.53 & 9.28 & 78.66 & \textbf{.3020} & 75.25 & \textbf{.2943} \\
\textbf{Linear $1/N_g$} & 29.59 & \textbf{4.65} & 7.22 & .0309 & .0606 & \textbf{4.06} & \textbf{8.03} & \textbf{79.40} & .2716 & \textbf{76.30} & .2635 \\
\midrule
\multicolumn{12}{c}{\textbf{(c) Clustering type (EQPO$_{ND}$, demographic labels masked)}} \\
\midrule
\textbf{No clustering} & \textbf{27.49} & 6.56 & \textbf{5.45} & .0448 & .0924 & 5.34 & 10.82 & 77.78 & .2704 & 73.84 & .2588 \\
\textbf{Random clustering} & 31.76 & 6.33 & 5.99 & .0389 & .0810 & 4.94 & 9.89 & 78.18 & .2870 & 74.50 & .2763 \\
\textbf{Reward-based (EQPO$_{ND}$)} & 30.59 & \textbf{5.70} & 5.78 & \textbf{.0356} & \textbf{.0715} & \textbf{3.91} & \textbf{7.85} & \textbf{79.20} & \textbf{.2944} & \textbf{76.22} & \textbf{.2843} \\
\bottomrule
\end{tabular}
}
\end{table*}

We ablate three components of EQPO on the QoQ-Med3-8B backbone: the cluster-count limit $k$, the form of the temperature scaling, and the choice of clustering procedure (Tab.~\ref{tab:ablations}).

\textbf{Cluster count.} With no clustering ($k = 1$, equivalent to per-domain weighting only), both fairness and accuracy degrade noticeably. Increasing the upper limit from $k = 10$ to $k = 20$ does not hurt performance because the elbow method automatically selects a smaller value when more clusters do not improve fit; this means the limit can be set generously without computational regret.

\textbf{Scaling form.} Linear scaling ($1/T_g$ without the square root) is theoretically optimal in the sense of inverse-frequency reweighting, but the resulting scaling factors span several orders of magnitude across groups and trigger numerical instability during gradient steps. Sublinear (square-root) scaling, used by default in EQPO, achieves the strongest fairness-utility trade-off in practice.

\textbf{Clustering type.} Reward-based clustering outperforms both no-clustering and random-clustering ablations on every fairness metric, confirming that the reward vectors carry task-specific difficulty information that random partitions cannot supply. Compared to random clustering, reward-based clustering reduces the maximum F1 gap by 11.7\% and accuracy standard deviation by 20.9\% while improving F1.

\section{Cluster Stability and Demographic Composition Details}
\label{app:cluster_stability}

This section provides the full cluster-stability and demographic-alignment analyses summarized in §\ref{sec:cluster_stability}. All numbers are from the EQPO$_{ND}$ run on QoQ-Med3-8B with demographic labels masked during training; ground-truth demographic labels are used post-hoc for the contingency analysis only.

\textbf{Cluster stability over training.} We measure cluster stability via Jaccard co-assignment similarity between consecutive training checkpoints. At each checkpoint, we identify the prompts present in both the current and previous evaluation and compute the Jaccard index over their pairwise co-assignment matrices; a value of $1.000$ indicates identical cluster assignments. The global average across all 7 domains is plotted as Fig.~\ref{fig:cluster_contingency}(c) of the main paper. Tab.~\ref{tab:jaccard_per_domain} shows the per-domain Jaccard at the end of training: six of seven domains have perfect Jaccard $1.000$, with CheXpert at $0.860$.

\begin{table}[h]
\centering
\caption{Per-domain Jaccard similarity at the final iteration.}
\label{tab:jaccard_per_domain}
\small
\begin{tabular}{ll}
\toprule
\textbf{Domain} & \textbf{Jaccard} \\
\midrule
COVID-BLUES & 1.000 \\
CheXpert & 0.860 \\
HAM10000 & 1.000 \\
Hemorrhage & 1.000 \\
ISIC-2020 & 1.000 \\
PAD-UFES-20 & 1.000 \\
VinDr-Mammo & 1.000 \\
\midrule
\textbf{Global avg.} & \textbf{0.980} \\
\bottomrule
\end{tabular}
\end{table}

\textbf{Cluster--demographic contingency.} The discovered clusters align meaningfully with held-out ground-truth demographics, even though demographic labels are masked at training time; we visualize this for VinDr-Mammo and HAM10000 in Fig.~\ref{fig:cluster_contingency} of the main paper.

\textbf{Mean scale per cluster.} Combining the contingencies with the EQPO$_{ND}$ scale assigned to each cluster (Tab.~\ref{tab:cluster_scale}) shows that the discovered minority clusters receive amplified learning signals (scale $\gg 1$) while the majority clusters are attenuated (scale $\approx 0$).

\begin{table}[h]
\centering
\caption{Mean EQPO$_{ND}$ scale per discovered cluster.}
\label{tab:cluster_scale}
\small
\begin{tabular}{llcl}
\toprule
\textbf{Domain} & \textbf{Cluster} & \textbf{Mean scale} & \textbf{Composition} \\
\midrule
HAM10000 & 0 & 2.113 & elderly minority \\
HAM10000 & 1 & 0.001 & majority \\
VinDr-Mammo & 0 & 0.936 & elderly minority \\
VinDr-Mammo & 1 & 0.001 & majority \\
\bottomrule
\end{tabular}
\end{table}

\section{LLM-as-a-Judge Reward Experiment}
\label{app:judge_reward}

To verify that EQPO's scaling mechanism behaves correctly under reward functions richer than binary correctness, we conducted a feasibility-check experiment in which a portion of the reward signal comes from an LLM-as-a-judge. Specifically, we use Gemini-3.1-Flash-Lite to score the reasoning quality of each rollout on a 1--5 scale and add the (normalized) judge score to the standard accuracy reward. We trained both GRPO+RS and EQPO on the QoQ-Med3-8B backbone for 50 steps under this reward and report the resulting metrics in Tab.~\ref{tab:judge_reward}.

\begin{table}[h]
\centering
\caption{\textbf{LLM-as-a-judge reward results on QoQ-Med3-8B (50 training steps).} Reward signal is supplied by Gemini-3.1-Flash-Lite scoring reasoning quality on a 1--5 scale, in addition to the binary accuracy reward used elsewhere in the paper. Lower is better for fairness metrics ($\downarrow$); higher is better for performance and combined metrics ($\uparrow$).}
\label{tab:judge_reward}
\resizebox{\textwidth}{!}{
\small
\setlength{\tabcolsep}{4.5pt}
\begin{tabular}{l|ccccccc|cc|cc}
\toprule
\textbf{Training Method} & \textbf{PP $\downarrow$} & \textbf{EOD $\downarrow$} & \textbf{FPR$_{\text{Diff}}$ $\downarrow$} & \textbf{$\sigma_{\text{F1}}$ $\downarrow$} & \textbf{$\Delta$F1 $\downarrow$} & \textbf{$\sigma_{\text{Acc}}$ $\downarrow$} & \textbf{$\Delta$Acc $\downarrow$} & \textbf{Acc $\uparrow$} & \textbf{F1 $\uparrow$} & \textbf{Acc$_{\text{ES}}$ $\uparrow$} & \textbf{F1$_{\text{ES}}$ $\uparrow$} \\
\midrule
\textbf{GRPO+RS} & \textbf{24.82} & 7.58 & 9.99 & .0369 & .0796 & 4.21 & 8.95 & 76.55 & .3028 & 73.46 & .2920 \\
\textbf{EQPO} & 25.16 & \textbf{6.01} & \textbf{9.57} & \textbf{.0299} & \textbf{.0661} & \textbf{4.12} & \textbf{8.59} & \textbf{77.79} & \textbf{.3056} & \textbf{74.71} & \textbf{.2967} \\
\bottomrule
\end{tabular}
}
\end{table}

EQPO improves over GRPO+RS on 10 of 11 metrics; the only exception is PP, where the two runs differ by $0.34$.

\section{Limitations}
\label{app:limitations}

EQPO has several limitations worth acknowledging. First, our evaluation focuses on predictive-parity-style fairness metrics, which are necessary but not sufficient for equitable clinical outcomes; a deployed system would also need to be validated on downstream measures such as treatment recommendation, time-to-diagnosis, and patient-reported outcomes. Second, when demographic labels are unavailable, EQPO$_{ND}$'s reward-based clusters may correlate with confounded attributes such as scanner type or institution rather than the demographic structure of interest. Future multi-institute studies could validate whether this is the case or not. Finally, our results are obtained on 7 public datasets covering 5 imaging modalities; the method has not been validated on time-series, electronic health record, or genomics modalities, and is intended as a research prototype rather than a deployment-ready system.

\section{Impact Statement}
\label{app:impact}

This paper presents work whose goal is to advance the field of Machine Learning, specifically addressing fairness in AI-assisted clinical diagnosis. We highlight several considerations relevant to the broader impact of this research.

Our work explicitly targets demographic disparities in healthcare AI, recognizing that biased systems can perpetuate existing inequalities. By developing EQPO, we aim to reduce performance gaps across age and gender groups, promoting more equitable diagnostic AI. We acknowledge that our demographic categorizations may not capture all dimensions of patient diversity, and future work should consider additional protected attributes and intersectional identities.

All experiments used publicly available, anonymized clinical datasets obtained in compliance with their respective licenses. No human subjects were directly involved, and no new clinical data was collected.

We emphasize that these models are research prototypes and should not be used for clinical decision-making without regulatory approval and clinical validation. Deployment of AI in healthcare requires careful consideration of local regulations, clinical workflows, and continuous monitoring for unintended consequences.

\section{The Use of Large Language Models (LLMs)}
We used ChatGPT for grammar corrections and debugging assistance, including explaining error messages and suggesting fixes. The model did not contribute research ideas, methods, experimental design, data, analyses or results. All changes were reviewed and implemented by the authors, who take full responsibility for the manuscript.

\begin{table*}[t]
\centering
\caption{\textbf{Fairness and performance metrics on Qwen-2.5-VL-7B}, in the same format as Tab.~\ref{tab:fairness_performance}. Lower is better for fairness metrics ($\downarrow$); higher is better for performance and combined metrics ($\uparrow$). Bold marks the best result in each column.}
\label{tab:qwen_main_result}
\resizebox{\textwidth}{!}{
\small
\setlength{\tabcolsep}{4.5pt}
\begin{tabular}{l|ccccccc|cc|cc}
\toprule
\multirow{2}{*}{\textbf{Training Method}} & \multicolumn{7}{c|}{\textbf{Fairness Metrics}} & \multicolumn{2}{c|}{\textbf{Perf. Metrics}} & \multicolumn{2}{c}{\textbf{Combined}} \\
\cmidrule(lr){2-8} \cmidrule(lr){9-10} \cmidrule(lr){11-12}
& \textbf{PP $\downarrow$} & \textbf{EOD $\downarrow$} & \textbf{FPR$_{\text{Diff}}$ $\downarrow$} & \textbf{$\sigma_{\text{F1}}$ $\downarrow$} & \textbf{$\Delta$F1 $\downarrow$} & \textbf{$\sigma_{\text{Acc}}$ $\downarrow$} & \textbf{$\Delta$Acc $\downarrow$} & \textbf{Acc $\uparrow$} & \textbf{F1 $\uparrow$} & \textbf{Acc$_{\text{ES}}$ $\uparrow$} & \textbf{F1$_{\text{ES}}$ $\uparrow$} \\
\midrule
\textbf{Re++~\citep{hu2025reinforce}} & 15.18 & 7.79 & 6.23 & .0322 & .0650 & 4.71 & 9.61 & 75.32 & .2612 & 71.93 & .2531 \\
\textbf{RLOO~\citep{rloo}} & 21.73 & 6.58 & 5.12 & .0326 & .0705 & 5.10 & 10.56 & 79.67 & .2479 & 75.80 & .2400 \\
\textbf{GRPO~\citep{shao2024deepseekmath}} & \textbf{11.39} & 9.09 & 4.61 & .0463 & .0973 & 4.68 & 9.43 & 80.45 & .2550 & 76.85 & .2437 \\
\textbf{GRPO+RS~\citep{puyol2021fairness}} & 21.56 & 8.09 & 4.96 & .0316 & .0636 & \textbf{3.97} & \textbf{8.11} & 73.99 & \textbf{.2657} & 70.57 & .2576 \\
\textbf{GRPO+DRO~\citep{sagawa2019distributionally}} & 14.51 & 7.41 & 7.42 & .0326 & .0654 & 5.62 & 11.50 & 75.10 & .2586 & 71.10 & .2504 \\
\midrule
\textbf{EQPO} & 16.80 & \textbf{5.55} & \textbf{4.39} & \textbf{.0229} & \textbf{.0452} & 4.41 & 8.93 & \textbf{80.75} & .2647 & \textbf{77.34} & \textbf{.2588} \\
\bottomrule
\end{tabular}
}
\end{table*}

\begin{table}[ht]
\centering
\caption{\textbf{Relative F1 score improvements (\%) for EQPO vs GRPO across demographic groups.} Values show the relative improvement ($\Delta\%$), GRPO baseline F1 score, and EQPO F1 score for each demographic group.}
\label{tab:group_scores_raw}
\small
\begin{tabular}{lcccccc}
\toprule
\multirow{2}{*}{\textbf{Group}} & \multicolumn{6}{c}{\textbf{Dataset}} \\
\cmidrule(lr){2-7}
 & \textbf{CheXpert} & \textbf{ISIC-2020} & \textbf{Hemorrhage} & \textbf{HAM10000} & \textbf{PAD-UFES-20} & \textbf{VinDr-Mammo} \\
\midrule
\textbf{a1} & & & & & & \\
$\Delta\%$ & +31.44 & -0.14 & -3.40 & -20.95 & 0.00 & -6.21 \\
GRPO & 0.318 & 0.495 & 0.721 & 0.383 & 0.462 & 0.243 \\
EQPO & 0.418 & 0.494 & 0.696 & 0.302 & 0.462 & 0.228 \\
\midrule
\textbf{a2} & & & & & & \\
$\Delta\%$ & +24.23 & -0.09 & +4.90 & -28.91 & +1.56 & +5.24 \\
GRPO & 0.296 & 0.496 & 0.600 & 0.262 & 0.385 & 0.234 \\
EQPO & 0.368 & 0.496 & 0.629 & 0.186 & 0.391 & 0.246 \\
\midrule
\textbf{a3} & & & & & & \\
$\Delta\%$ & +33.18 & +1.65 & +18.77 & +39.18 & +3.71 & +11.29 \\
GRPO & 0.283 & 0.564 & 0.679 & 0.222 & 0.190 & 0.195 \\
EQPO & 0.377 & 0.574 & 0.806 & 0.309 & 0.197 & 0.217 \\
\midrule
\textbf{a4} & & & & & & \\
$\Delta\%$ & +21.60 & +27.08 & -- & +6.03 & +6.22 & -13.85 \\
GRPO & 0.302 & 0.469 & -- & 0.185 & 0.221 & 0.238 \\
EQPO & 0.368 & 0.595 & -- & 0.196 & 0.234 & 0.205 \\
\midrule
\textbf{Female} & & & & & & \\
$\Delta\%$ & +24.45 & +6.90 & +3.54 & +19.21 & +4.20 & -- \\
GRPO & 0.320 & 0.517 & 0.773 & 0.262 & 0.247 & -- \\
EQPO & 0.398 & 0.553 & 0.800 & 0.313 & 0.258 & -- \\
\midrule
\textbf{Male} & & & & & & \\
$\Delta\%$ & +34.35 & +2.26 & +6.97 & +9.52 & +2.67 & -- \\
GRPO & 0.253 & 0.546 & 0.628 & 0.240 & 0.214 & -- \\
EQPO & 0.340 & 0.558 & 0.672 & 0.263 & 0.220 & -- \\
\midrule
\textbf{Average $\Delta\%$} & +28.21 & +6.28 & +6.16 & +4.01 & +3.06 & -0.88 \\
\bottomrule
\end{tabular}
\end{table}

\begin{table*}[t]
\centering
\caption{\textbf{Detailed fairness and performance metrics per dataset and demographic group for Reinforce++ on Qwen-2.5-VL.} Results shown for both age groups (a1-a4) and gender groups across all evaluation datasets. Higher values are better for accuracy, TPR, and F1; lower values are better for FPR and FDR.}
\label{tab:detailed_demographics_repp_qwen}
\vspace{-0.5em}
\resizebox{\textwidth}{!}{
\small
\setlength{\tabcolsep}{2.5pt}
\begin{tabular}{ll|cccc|cccc|cccc}
\toprule
\multirow{2}{*}{\textbf{Dataset}} & \multirow{2}{*}{\textbf{Group}} & \multicolumn{4}{c|}{\textbf{Performance Metrics}} & \multicolumn{4}{c|}{\textbf{Fairness Metrics}} & \multicolumn{4}{c}{\textbf{Disparity Metrics}} \\
\cmidrule(lr){3-6} \cmidrule(lr){7-10} \cmidrule(lr){11-14}
& & \textbf{Acc} & \textbf{F1} & \textbf{TPR} & \textbf{FPR} & \textbf{FDR} & \textbf{$\sigma_{\text{Acc}}$} & \textbf{$\sigma_{\text{F1}}$} & \textbf{$\sigma_{\text{TPR}}$} & \textbf{$\Delta$Acc} & \textbf{$\Delta$F1} & \textbf{$\Delta$TPR} & \textbf{$\Delta$FPR} \\
\midrule
\multicolumn{14}{c}{\textbf{Age Groups}} \\
\midrule
\multirow{4}{*}{\textbf{ChexPert}} 
& a1 & .833 & .130 & .138 & .064 & .158 & \multirow{4}{*}{.076} & \multirow{4}{*}{.012} & \multirow{4}{*}{.018} & \multirow{4}{*}{.184} & \multirow{4}{*}{.028} & \multirow{4}{*}{.038} & \multirow{4}{*}{.009} \\
& a2 & .748 & .102 & .118 & .068 & .139 & & & & & & & \\
& a3 & .770 & .120 & .114 & .070 & .202 & & & & & & & \\
& a4 & .649 & .125 & .151 & .074 & .223 & & & & & & & \\
\midrule
\multirow{4}{*}{\textbf{HAM10000}} 
& a1 & .824 & .347 & .426 & .252 & .317 & \multirow{4}{*}{.077} & \multirow{4}{*}{.094} & \multirow{4}{*}{.099} & \multirow{4}{*}{.183} & \multirow{4}{*}{.225} & \multirow{4}{*}{.218} & \multirow{4}{*}{.068} \\
\bottomrule
\end{tabular}
}
\vspace{-1em}
\end{table*}
\begin{table*}[t]
\centering
\caption{\textbf{Detailed fairness and performance metrics per dataset and demographic group for RLOO on Qwen-2.5-VL.} Results shown for both age groups (a1-a4) and gender groups across all evaluation datasets. Higher values are better for accuracy, TPR, and F1; lower values are better for FPR and FDR.}
\label{tab:detailed_demographics_rloo_qwen}
\vspace{-0.5em}
\resizebox{\textwidth}{!}{
\small
\setlength{\tabcolsep}{2.5pt}
\begin{tabular}{ll|cccc|cccc|cccc}
\toprule
\multirow{2}{*}{\textbf{Dataset}} & \multirow{2}{*}{\textbf{Group}} & \multicolumn{4}{c|}{\textbf{Performance Metrics}} & \multicolumn{4}{c|}{\textbf{Fairness Metrics}} & \multicolumn{4}{c}{\textbf{Disparity Metrics}} \\
\cmidrule(lr){3-6} \cmidrule(lr){7-10} \cmidrule(lr){11-14}
& & \textbf{Acc} & \textbf{F1} & \textbf{TPR} & \textbf{FPR} & \textbf{FDR} & \textbf{$\sigma_{\text{Acc}}$} & \textbf{$\sigma_{\text{F1}}$} & \textbf{$\sigma_{\text{TPR}}$} & \textbf{$\Delta$Acc} & \textbf{$\Delta$F1} & \textbf{$\Delta$TPR} & \textbf{$\Delta$FPR} \\
\midrule
\multicolumn{14}{c}{\textbf{Age Groups}} \\
\midrule
\multirow{4}{*}{\textbf{ChexPert}} 
& a1 & .833 & .285 & .338 & .096 & .221 & \multirow{4}{*}{.066} & \multirow{4}{*}{.067} & \multirow{4}{*}{.083} & \multirow{4}{*}{.161} & \multirow{4}{*}{.149} & \multirow{4}{*}{.186} & \multirow{4}{*}{.080} \\
& a2 & .746 & .136 & .152 & .125 & .201 & & & & & & & \\
& a3 & .767 & .154 & .175 & .142 & .449 & & & & & & & \\
& a4 & .673 & .179 & .200 & .176 & .114 & & & & & & & \\
\midrule
\multirow{4}{*}{\textbf{HAM10000}} 
& a1 & .924 & .314 & .327 & .309 & .364 & \multirow{4}{*}{.114} & \multirow{4}{*}{.087} & \multirow{4}{*}{.051} & \multirow{4}{*}{.243} & \multirow{4}{*}{.199} & \multirow{4}{*}{.107} & \multirow{4}{*}{.116} \\
\bottomrule
\end{tabular}
}
\vspace{-1em}
\end{table*}
\begin{table*}[t]
\centering
\caption{\textbf{Detailed fairness and performance metrics per dataset and demographic group for GRPO on Qwen-2.5-VL.} Results shown for both age groups (a1-a4) and gender groups across all evaluation datasets. Higher values are better for accuracy, TPR, and F1; lower values are better for FPR and FDR.}
\label{tab:detailed_demographics_grpo_qwen}
\vspace{-0.5em}
\resizebox{\textwidth}{!}{
\small
\setlength{\tabcolsep}{2.5pt}
\begin{tabular}{ll|cccc|cccc|cccc}
\toprule
\multirow{2}{*}{\textbf{Dataset}} & \multirow{2}{*}{\textbf{Group}} & \multicolumn{4}{c|}{\textbf{Performance Metrics}} & \multicolumn{4}{c|}{\textbf{Fairness Metrics}} & \multicolumn{4}{c}{\textbf{Disparity Metrics}} \\
\cmidrule(lr){3-6} \cmidrule(lr){7-10} \cmidrule(lr){11-14}
& & \textbf{Acc} & \textbf{F1} & \textbf{TPR} & \textbf{FPR} & \textbf{FDR} & \textbf{$\sigma_{\text{Acc}}$} & \textbf{$\sigma_{\text{F1}}$} & \textbf{$\sigma_{\text{TPR}}$} & \textbf{$\Delta$Acc} & \textbf{$\Delta$F1} & \textbf{$\Delta$TPR} & \textbf{$\Delta$FPR} \\
\midrule
\multicolumn{14}{c}{\textbf{Age Groups}} \\
\midrule
\multirow{4}{*}{\textbf{ChexPert}} 
& a1 & .807 & .235 & .338 & .133 & .188 & \multirow{4}{*}{.049} & \multirow{4}{*}{.041} & \multirow{4}{*}{.062} & \multirow{4}{*}{.112} & \multirow{4}{*}{.091} & \multirow{4}{*}{.141} & \multirow{4}{*}{.051} \\
& a2 & .766 & .192 & .228 & .136 & .160 & & & & & & & \\
& a3 & .785 & .163 & .196 & .141 & .160 & & & & & & & \\
& a4 & .695 & .254 & .282 & .184 & .096 & & & & & & & \\
\midrule
\multirow{4}{*}{\textbf{HAM10000}} 
& a1 & .936 & .317 & .333 & .303 & .031 & \multirow{4}{*}{.116} & \multirow{4}{*}{.076} & \multirow{4}{*}{.059} & \multirow{4}{*}{.239} & \multirow{4}{*}{.168} & \multirow{4}{*}{.134} & \multirow{4}{*}{.112} \\
\bottomrule
\end{tabular}
}
\vspace{-1em}
\end{table*}
\begin{table*}[t]
\centering
\caption{\textbf{Detailed fairness and performance metrics per dataset and demographic group for GRPO with Resampling on Qwen-2.5-VL.} Results shown for both age groups (a1-a4) and gender groups across all evaluation datasets. Higher values are better for accuracy, TPR, and F1; lower values are better for FPR and FDR.}
\label{tab:detailed_demographics_rs_qwen}
\vspace{-0.5em}
\resizebox{\textwidth}{!}{
\small
\setlength{\tabcolsep}{2.5pt}
\begin{tabular}{ll|cccc|cccc|cccc}
\toprule
\multirow{2}{*}{\textbf{Dataset}} & \multirow{2}{*}{\textbf{Group}} & \multicolumn{4}{c|}{\textbf{Performance Metrics}} & \multicolumn{4}{c|}{\textbf{Fairness Metrics}} & \multicolumn{4}{c}{\textbf{Disparity Metrics}} \\
\cmidrule(lr){3-6} \cmidrule(lr){7-10} \cmidrule(lr){11-14}
& & \textbf{Acc} & \textbf{F1} & \textbf{TPR} & \textbf{FPR} & \textbf{FDR} & \textbf{$\sigma_{\text{Acc}}$} & \textbf{$\sigma_{\text{F1}}$} & \textbf{$\sigma_{\text{TPR}}$} & \textbf{$\Delta$Acc} & \textbf{$\Delta$F1} & \textbf{$\Delta$TPR} & \textbf{$\Delta$FPR} \\
\midrule
\multicolumn{14}{c}{\textbf{Age Groups}} \\
\midrule
\multirow{4}{*}{\textbf{ChexPert}} 
& a1 & .847 & .142 & .150 & .056 & .153 & \multirow{4}{*}{.083} & \multirow{4}{*}{.018} & \multirow{4}{*}{.029} & \multirow{4}{*}{.203} & \multirow{4}{*}{.044} & \multirow{4}{*}{.059} & \multirow{4}{*}{.022} \\
& a2 & .762 & .125 & .153 & .056 & .079 & & & & & & & \\
& a3 & .754 & .098 & .094 & .078 & .279 & & & & & & & \\
& a4 & .644 & .124 & .153 & .076 & .321 & & & & & & & \\
\midrule
\multirow{4}{*}{\textbf{HAM10000}} 
& a1 & .785 & .307 & .354 & .302 & .341 & \multirow{4}{*}{.057} & \multirow{4}{*}{.071} & \multirow{4}{*}{.070} & \multirow{4}{*}{.138} & \multirow{4}{*}{.158} & \multirow{4}{*}{.160} & \multirow{4}{*}{.111} \\
\bottomrule
\end{tabular}
}
\vspace{-1em}
\end{table*}
\begin{table*}[t]
\centering
\caption{\textbf{Detailed fairness and performance metrics per dataset and demographic group for GRPO with Group DRO on Qwen-2.5-VL.} Results shown for both age groups (a1-a4) and gender groups across all evaluation datasets. Higher values are better for accuracy, TPR, and F1; lower values are better for FPR and FDR.}
\label{tab:detailed_demographics_dro_qwen}
\vspace{-0.5em}
\resizebox{\textwidth}{!}{
\small
\setlength{\tabcolsep}{2.5pt}
\begin{tabular}{ll|cccc|cccc|cccc}
\toprule
\multirow{2}{*}{\textbf{Dataset}} & \multirow{2}{*}{\textbf{Group}} & \multicolumn{4}{c|}{\textbf{Performance Metrics}} & \multicolumn{4}{c|}{\textbf{Fairness Metrics}} & \multicolumn{4}{c}{\textbf{Disparity Metrics}} \\
\cmidrule(lr){3-6} \cmidrule(lr){7-10} \cmidrule(lr){11-14}
& & \textbf{Acc} & \textbf{F1} & \textbf{TPR} & \textbf{FPR} & \textbf{FDR} & \textbf{$\sigma_{\text{Acc}}$} & \textbf{$\sigma_{\text{F1}}$} & \textbf{$\sigma_{\text{TPR}}$} & \textbf{$\Delta$Acc} & \textbf{$\Delta$F1} & \textbf{$\Delta$TPR} & \textbf{$\Delta$FPR} \\
\midrule
\multicolumn{14}{c}{\textbf{Age Groups}} \\
\midrule
\multirow{4}{*}{\textbf{ChexPert}} 
& a1 & .847 & .142 & .150 & .056 & .153 & \multirow{4}{*}{.092} & \multirow{4}{*}{.021} & \multirow{4}{*}{.014} & \multirow{4}{*}{.221} & \multirow{4}{*}{.049} & \multirow{4}{*}{.033} & \multirow{4}{*}{.031} \\
& a2 & .754 & .105 & .132 & .063 & .127 & & & & & & & \\
& a3 & .767 & .115 & .117 & .070 & .124 & & & & & & & \\
& a4 & .625 & .092 & .132 & .087 & .245 & & & & & & & \\
\midrule
\multirow{4}{*}{\textbf{HAM10000}} 
& a1 & .821 & .327 & .374 & .283 & .333 & \multirow{4}{*}{.059} & \multirow{4}{*}{.070} & \multirow{4}{*}{.076} & \multirow{4}{*}{.131} & \multirow{4}{*}{.158} & \multirow{4}{*}{.180} & \multirow{4}{*}{.097} \\
\bottomrule
\end{tabular}
}
\vspace{-1em}
\end{table*}
\begin{table*}[t]
\centering
\caption{\textbf{Detailed fairness and performance metrics per dataset and demographic group for EQPO on Qwen-2.5-VL.} Results shown for both age groups (a1-a4) and gender groups across all evaluation datasets. Higher values are better for accuracy, TPR, and F1; lower values are better for FPR and FDR.}
\label{tab:detailed_demographics_eqpo_qwen}
\vspace{-0.5em}
\resizebox{\textwidth}{!}{
\small
\setlength{\tabcolsep}{2.5pt}
\begin{tabular}{ll|cccc|cccc|cccc}
\toprule
\multirow{2}{*}{\textbf{Dataset}} & \multirow{2}{*}{\textbf{Group}} & \multicolumn{4}{c|}{\textbf{Performance Metrics}} & \multicolumn{4}{c|}{\textbf{Fairness Metrics}} & \multicolumn{4}{c}{\textbf{Disparity Metrics}} \\
\cmidrule(lr){3-6} \cmidrule(lr){7-10} \cmidrule(lr){11-14}
& & \textbf{Acc} & \textbf{F1} & \textbf{TPR} & \textbf{FPR} & \textbf{FDR} & \textbf{$\sigma_{\text{Acc}}$} & \textbf{$\sigma_{\text{F1}}$} & \textbf{$\sigma_{\text{TPR}}$} & \textbf{$\Delta$Acc} & \textbf{$\Delta$F1} & \textbf{$\Delta$TPR} & \textbf{$\Delta$FPR} \\
\midrule
\multicolumn{14}{c}{\textbf{Age Groups}} \\
\midrule
\multirow{4}{*}{\textbf{ChexPert}} 
& a1 & .813 & .161 & .225 & .109 & .140 & \multirow{4}{*}{.063} & \multirow{4}{*}{.015} & \multirow{4}{*}{.030} & \multirow{4}{*}{.142} & \multirow{4}{*}{.031} & \multirow{4}{*}{.065} & \multirow{4}{*}{.068} \\
& a2 & .771 & .149 & .166 & .096 & .104 & & & & & & & \\
& a3 & .792 & .132 & .160 & .118 & .105 & & & & & & & \\
& a4 & .671 & .130 & .177 & .164 & .076 & & & & & & & \\
\midrule
\multirow{4}{*}{\textbf{HAM10000}} 
& a1 & .915 & .304 & .300 & .212 & .024 & \multirow{4}{*}{.089} & \multirow{4}{*}{.045} & \multirow{4}{*}{.075} & \multirow{4}{*}{.183} & \multirow{4}{*}{.096} & \multirow{4}{*}{.164} & \multirow{4}{*}{.069} \\
\bottomrule
\end{tabular}
}
\vspace{-1em}
\end{table*}

\begin{table*}[t]
\centering
\caption{\textbf{Detailed fairness and performance metrics per dataset and demographic group for Reinforce++ on MedGemma.} Results shown for both age groups (a1-a4) and gender groups across all evaluation datasets. Higher values are better for accuracy, TPR, and F1; lower values are better for FPR and FDR.}
\label{tab:detailed_demographics_repp_medgemma}
\vspace{-0.5em}
\resizebox{\textwidth}{!}{
\small
\setlength{\tabcolsep}{2.5pt}
\begin{tabular}{ll|cccc|cccc|cccc}
\toprule
\multirow{2}{*}{\textbf{Dataset}} & \multirow{2}{*}{\textbf{Group}} & \multicolumn{4}{c|}{\textbf{Performance Metrics}} & \multicolumn{4}{c|}{\textbf{Fairness Metrics}} & \multicolumn{4}{c}{\textbf{Disparity Metrics}} \\
\cmidrule(lr){3-6} \cmidrule(lr){7-10} \cmidrule(lr){11-14}
& & \textbf{Acc} & \textbf{F1} & \textbf{TPR} & \textbf{FPR} & \textbf{FDR} & \textbf{$\sigma_{\text{Acc}}$} & \textbf{$\sigma_{\text{F1}}$} & \textbf{$\sigma_{\text{TPR}}$} & \textbf{$\Delta$Acc} & \textbf{$\Delta$F1} & \textbf{$\Delta$TPR} & \textbf{$\Delta$FPR} \\
\midrule
\multicolumn{14}{c}{\textbf{Age Groups}} \\
\midrule
\multirow{4}{*}{\textbf{ChexPert}} 
& a1 & .793 & .288 & .338 & .173 & .643 & \multirow{4}{*}{.038} & \multirow{4}{*}{.025} & \multirow{4}{*}{.031} & \multirow{4}{*}{.092} & \multirow{4}{*}{.056} & \multirow{4}{*}{.063} & \multirow{4}{*}{.039} \\
& a2 & .745 & .260 & .299 & .161 & .659 & & & & & & & \\
& a3 & .761 & .269 & .298 & .161 & .647 & & & & & & & \\
& a4 & .702 & .316 & .361 & .134 & .483 & & & & & & & \\
\midrule
\multirow{4}{*}{\textbf{HAM10000}} 
& a1 & .927 & .312 & .323 & .303 & .031 & \multirow{4}{*}{.107} & \multirow{4}{*}{.083} & \multirow{4}{*}{.054} & \multirow{4}{*}{.223} & \multirow{4}{*}{.197} & \multirow{4}{*}{.124} & \multirow{4}{*}{.138} \\
\bottomrule
\end{tabular}
}
\vspace{-1em}
\end{table*}
\begin{table*}[t]
\centering
\caption{\textbf{Detailed fairness and performance metrics per dataset and demographic group for RLOO on MedGemma.} Results shown for both age groups (a1-a4) and gender groups across all evaluation datasets. Higher values are better for accuracy, TPR, and F1; lower values are better for FPR and FDR.}
\label{tab:detailed_demographics_rloo_medgemma}
\vspace{-0.5em}
\resizebox{\textwidth}{!}{
\small
\setlength{\tabcolsep}{2.5pt}
\begin{tabular}{ll|cccc|cccc|cccc}
\toprule
\multirow{2}{*}{\textbf{Dataset}} & \multirow{2}{*}{\textbf{Group}} & \multicolumn{4}{c|}{\textbf{Performance Metrics}} & \multicolumn{4}{c|}{\textbf{Fairness Metrics}} & \multicolumn{4}{c}{\textbf{Disparity Metrics}} \\
\cmidrule(lr){3-6} \cmidrule(lr){7-10} \cmidrule(lr){11-14}
& & \textbf{Acc} & \textbf{F1} & \textbf{TPR} & \textbf{FPR} & \textbf{FDR} & \textbf{$\sigma_{\text{Acc}}$} & \textbf{$\sigma_{\text{F1}}$} & \textbf{$\sigma_{\text{TPR}}$} & \textbf{$\Delta$Acc} & \textbf{$\Delta$F1} & \textbf{$\Delta$TPR} & \textbf{$\Delta$FPR} \\
\midrule
\multicolumn{14}{c}{\textbf{Age Groups}} \\
\midrule
\multirow{4}{*}{\textbf{ChexPert}} 
& a1 & .900 & .533 & .600 & .082 & .185 & \multirow{4}{*}{.077} & \multirow{4}{*}{.091} & \multirow{4}{*}{.090} & \multirow{4}{*}{.189} & \multirow{4}{*}{.193} & \multirow{4}{*}{.196} & \multirow{4}{*}{.100} \\
& a2 & .817 & .363 & .429 & .110 & .241 & & & & & & & \\
& a3 & .810 & .351 & .404 & .122 & .307 & & & & & & & \\
& a4 & .711 & .339 & .432 & .183 & .378 & & & & & & & \\
\midrule
\multirow{4}{*}{\textbf{HAM10000}} 
& a1 & .933 & .316 & .333 & .333 & .033 & \multirow{4}{*}{.110} & \multirow{4}{*}{.065} & \multirow{4}{*}{.058} & \multirow{4}{*}{.228} & \multirow{4}{*}{.139} & \multirow{4}{*}{.138} & \multirow{4}{*}{.147} \\
\bottomrule
\end{tabular}
}
\vspace{-1em}
\end{table*}
\begin{table*}[t]
\centering
\caption{\textbf{Detailed fairness and performance metrics per dataset and demographic group for GRPO on MedGemma.} Results shown for both age groups (a1-a4) and gender groups across all evaluation datasets. Higher values are better for accuracy, TPR, and F1; lower values are better for FPR and FDR.}
\label{tab:detailed_demographics_grpo_medgemma}
\vspace{-0.5em}
\resizebox{\textwidth}{!}{
\small
\setlength{\tabcolsep}{2.5pt}
\begin{tabular}{ll|cccc|cccc|cccc}
\toprule
\multirow{2}{*}{\textbf{Dataset}} & \multirow{2}{*}{\textbf{Group}} & \multicolumn{4}{c|}{\textbf{Performance Metrics}} & \multicolumn{4}{c|}{\textbf{Fairness Metrics}} & \multicolumn{4}{c}{\textbf{Disparity Metrics}} \\
\cmidrule(lr){3-6} \cmidrule(lr){7-10} \cmidrule(lr){11-14}
& & \textbf{Acc} & \textbf{F1} & \textbf{TPR} & \textbf{FPR} & \textbf{FDR} & \textbf{$\sigma_{\text{Acc}}$} & \textbf{$\sigma_{\text{F1}}$} & \textbf{$\sigma_{\text{TPR}}$} & \textbf{$\Delta$Acc} & \textbf{$\Delta$F1} & \textbf{$\Delta$TPR} & \textbf{$\Delta$FPR} \\
\midrule
\multicolumn{14}{c}{\textbf{Age Groups}} \\
\midrule
\multirow{4}{*}{\textbf{ChexPert}} 
& a1 & .893 & .318 & .342 & .038 & .092 & \multirow{4}{*}{.084} & \multirow{4}{*}{.015} & \multirow{4}{*}{.040} & \multirow{4}{*}{.202} & \multirow{4}{*}{.035} & \multirow{4}{*}{.093} & \multirow{4}{*}{.050} \\
& a2 & .814 & .296 & .279 & .059 & .174 & & & & & & & \\
& a3 & .824 & .283 & .248 & .052 & .159 & & & & & & & \\
& a4 & .691 & .302 & .306 & .088 & .211 & & & & & & & \\
\midrule
\multirow{4}{*}{\textbf{HAM10000}} 
& a1 & .918 & .383 & .369 & .279 & .252 & \multirow{4}{*}{.108} & \multirow{4}{*}{.086} & \multirow{4}{*}{.057} & \multirow{4}{*}{.233} & \multirow{4}{*}{.198} & \multirow{4}{*}{.120} & \multirow{4}{*}{.103} \\
\bottomrule
\end{tabular}
}
\vspace{-1em}
\end{table*}
\begin{table*}[t]
\centering
\caption{\textbf{Detailed fairness and performance metrics per dataset and demographic group for GRPO with Resampling on MedGemma.} Results shown for both age groups (a1-a4) and gender groups across all evaluation datasets. Higher values are better for accuracy, TPR, and F1; lower values are better for FPR and FDR.}
\label{tab:detailed_demographics_rs_medgemma}
\vspace{-0.5em}
\resizebox{\textwidth}{!}{
\small
\setlength{\tabcolsep}{2.5pt}
\begin{tabular}{ll|cccc|cccc|cccc}
\toprule
\multirow{2}{*}{\textbf{Dataset}} & \multirow{2}{*}{\textbf{Group}} & \multicolumn{4}{c|}{\textbf{Performance Metrics}} & \multicolumn{4}{c|}{\textbf{Fairness Metrics}} & \multicolumn{4}{c}{\textbf{Disparity Metrics}} \\
\cmidrule(lr){3-6} \cmidrule(lr){7-10} \cmidrule(lr){11-14}
& & \textbf{Acc} & \textbf{F1} & \textbf{TPR} & \textbf{FPR} & \textbf{FDR} & \textbf{$\sigma_{\text{Acc}}$} & \textbf{$\sigma_{\text{F1}}$} & \textbf{$\sigma_{\text{TPR}}$} & \textbf{$\Delta$Acc} & \textbf{$\Delta$F1} & \textbf{$\Delta$TPR} & \textbf{$\Delta$FPR} \\
\midrule
\multicolumn{14}{c}{\textbf{Age Groups}} \\
\midrule
\multirow{4}{*}{\textbf{ChexPert}} 
& a1 & .913 & .466 & .500 & .045 & .061 & \multirow{4}{*}{.072} & \multirow{4}{*}{.066} & \multirow{4}{*}{.066} & \multirow{4}{*}{.175} & \multirow{4}{*}{.147} & \multirow{4}{*}{.157} & \multirow{4}{*}{.061} \\
& a2 & .832 & .349 & .389 & .072 & .202 & & & & & & & \\
& a3 & .828 & .319 & .343 & .074 & .284 & & & & & & & \\
& a4 & .738 & .343 & .400 & .106 & .272 & & & & & & & \\
\midrule
\multirow{4}{*}{\textbf{HAM10000}} 
& a1 & .942 & .320 & .333 & .242 & .025 & \multirow{4}{*}{.111} & \multirow{4}{*}{.074} & \multirow{4}{*}{.037} & \multirow{4}{*}{.236} & \multirow{4}{*}{.170} & \multirow{4}{*}{.084} & \multirow{4}{*}{.070} \\
\bottomrule
\end{tabular}
}
\vspace{-1em}
\end{table*}

\begin{table*}[t]
\centering
\caption{\textbf{Detailed fairness and performance metrics per dataset and demographic group for GRPO with Group DRO on MedGemma.} Results shown for both age groups (a1-a4) and gender groups across all evaluation datasets. Higher values are better for accuracy, TPR, and F1; lower values are better for FPR and FDR.}
\label{tab:detailed_demographics_dro_medgemma}
\vspace{-0.5em}
\resizebox{\textwidth}{!}{
\small
\setlength{\tabcolsep}{2.5pt}
\begin{tabular}{ll|cccc|cccc|cccc}
\toprule
\multirow{2}{*}{\textbf{Dataset}} & \multirow{2}{*}{\textbf{Group}} & \multicolumn{4}{c|}{\textbf{Performance Metrics}} & \multicolumn{4}{c|}{\textbf{Fairness Metrics}} & \multicolumn{4}{c}{\textbf{Disparity Metrics}} \\
\cmidrule(lr){3-6} \cmidrule(lr){7-10} \cmidrule(lr){11-14}
& & \textbf{Acc} & \textbf{F1} & \textbf{TPR} & \textbf{FPR} & \textbf{FDR} & \textbf{$\sigma_{\text{Acc}}$} & \textbf{$\sigma_{\text{F1}}$} & \textbf{$\sigma_{\text{TPR}}$} & \textbf{$\Delta$Acc} & \textbf{$\Delta$F1} & \textbf{$\Delta$TPR} & \textbf{$\Delta$FPR} \\
\midrule
\multicolumn{14}{c}{\textbf{Age Groups}} \\
\midrule
\multirow{4}{*}{\textbf{ChexPert}} 
& a1 & .913 & .380 & .400 & .030 & .136 & \multirow{4}{*}{.064} & \multirow{4}{*}{.029} & \multirow{4}{*}{.032} & \multirow{4}{*}{.157} & \multirow{4}{*}{.060} & \multirow{4}{*}{.070} & \multirow{4}{*}{.062} \\
& a2 & .848 & .390 & .402 & .049 & .168 & & & & & & & \\
& a3 & .836 & .330 & .341 & .065 & .228 & & & & & & & \\
& a4 & .756 & .390 & .411 & .093 & .225 & & & & & & & \\
\midrule
\multirow{4}{*}{\textbf{HAM10000}} 
& a1 & .945 & .509 & .467 & .273 & .028 & \multirow{4}{*}{.116} & \multirow{4}{*}{.162} & \multirow{4}{*}{.116} & \multirow{4}{*}{.245} & \multirow{4}{*}{.373} & \multirow{4}{*}{.246} & \multirow{4}{*}{.088} \\
\bottomrule
\end{tabular}
}
\vspace{-1em}
\end{table*}
\begin{table*}[t]
\centering
\caption{\textbf{Detailed fairness and performance metrics per dataset and demographic group for EQPO$_{ND}$ on MedGemma.} Results shown for both age groups (a1-a4) and gender groups across all evaluation datasets. Higher values are better for accuracy, TPR, and F1; lower values are better for FPR and FDR.}
\label{tab:detailed_demographics_eqpo_nd_medgemma}
\vspace{-0.5em}
\resizebox{\textwidth}{!}{
\small
\setlength{\tabcolsep}{2.5pt}
\begin{tabular}{ll|cccc|cccc|cccc}
\toprule
\multirow{2}{*}{\textbf{Dataset}} & \multirow{2}{*}{\textbf{Group}} & \multicolumn{4}{c|}{\textbf{Performance Metrics}} & \multicolumn{4}{c|}{\textbf{Fairness Metrics}} & \multicolumn{4}{c}{\textbf{Disparity Metrics}} \\
\cmidrule(lr){3-6} \cmidrule(lr){7-10} \cmidrule(lr){11-14}
& & \textbf{Acc} & \textbf{F1} & \textbf{TPR} & \textbf{FPR} & \textbf{FDR} & \textbf{$\sigma_{\text{Acc}}$} & \textbf{$\sigma_{\text{F1}}$} & \textbf{$\sigma_{\text{TPR}}$} & \textbf{$\Delta$Acc} & \textbf{$\Delta$F1} & \textbf{$\Delta$TPR} & \textbf{$\Delta$FPR} \\
\midrule
\multicolumn{14}{c}{\textbf{Age Groups}} \\
\midrule
\multirow{4}{*}{\textbf{ChexPert}} 
& a1 & .860 & .342 & .475 & .096 & .207 & \multirow{4}{*}{.057} & \multirow{4}{*}{.037} & \multirow{4}{*}{.044} & \multirow{4}{*}{.138} & \multirow{4}{*}{.087} & \multirow{4}{*}{.088} & \multirow{4}{*}{.101} \\
& a2 & .800 & .366 & .387 & .146 & .523 & & & & & & & \\
& a3 & .799 & .379 & .390 & .162 & .544 & & & & & & & \\
& a4 & .722 & .430 & .449 & .198 & .393 & & & & & & & \\
\midrule
\multirow{4}{*}{\textbf{HAM10000}} 
& a1 & .897 & .301 & .296 & .255 & .694 & \multirow{4}{*}{.077} & \multirow{4}{*}{.035} & \multirow{4}{*}{.029} & \multirow{4}{*}{.163} & \multirow{4}{*}{.084} & \multirow{4}{*}{.071} & \multirow{4}{*}{.121} \\
\bottomrule
\end{tabular}
}
\vspace{-1em}
\end{table*}
\begin{table*}[t]
\centering
\caption{\textbf{Detailed fairness and performance metrics per dataset and demographic group for EQPO on MedGemma.} Results shown for both age groups (a1-a4) and gender groups across all evaluation datasets. Higher values are better for accuracy, TPR, and F1; lower values are better for FPR and FDR.}
\label{tab:detailed_demographics_eqpo_medgemma}
\vspace{-0.5em}
\resizebox{\textwidth}{!}{
\small
\setlength{\tabcolsep}{2.5pt}
\begin{tabular}{ll|cccc|cccc|cccc}
\toprule
\multirow{2}{*}{\textbf{Dataset}} & \multirow{2}{*}{\textbf{Group}} & \multicolumn{4}{c|}{\textbf{Performance Metrics}} & \multicolumn{4}{c|}{\textbf{Fairness Metrics}} & \multicolumn{4}{c}{\textbf{Disparity Metrics}} \\
\cmidrule(lr){3-6} \cmidrule(lr){7-10} \cmidrule(lr){11-14}
& & \textbf{Acc} & \textbf{F1} & \textbf{TPR} & \textbf{FPR} & \textbf{FDR} & \textbf{$\sigma_{\text{Acc}}$} & \textbf{$\sigma_{\text{F1}}$} & \textbf{$\sigma_{\text{TPR}}$} & \textbf{$\Delta$Acc} & \textbf{$\Delta$F1} & \textbf{$\Delta$TPR} & \textbf{$\Delta$FPR} \\
\midrule
\multicolumn{14}{c}{\textbf{Age Groups}} \\
\midrule
\multirow{4}{*}{\textbf{ChexPert}} 
& a1 & .900 & .359 & .388 & .045 & .063 & \multirow{4}{*}{.062} & \multirow{4}{*}{.019} & \multirow{4}{*}{.035} & \multirow{4}{*}{.151} & \multirow{4}{*}{.047} & \multirow{4}{*}{.076} & \multirow{4}{*}{.056} \\
& a2 & .828 & .354 & .351 & .063 & .224 & & & & & & & \\
& a3 & .833 & .328 & .330 & .065 & .239 & & & & & & & \\
& a4 & .749 & .375 & .406 & .101 & .332 & & & & & & & \\
\midrule
\multirow{4}{*}{\textbf{HAM10000}} 
& a1 & .933 & .315 & .327 & .273 & .028 & \multirow{4}{*}{.114} & \multirow{4}{*}{.074} & \multirow{4}{*}{.043} & \multirow{4}{*}{.238} & \multirow{4}{*}{.171} & \multirow{4}{*}{.088} & \multirow{4}{*}{.083} \\
\bottomrule
\end{tabular}
}
\vspace{-1em}
\end{table*}
\begin{table*}[t]
\centering
\caption{\textbf{RQ1: Fairness and performance metrics comparison against RL and fairness mitigation baselines.} For fairness metrics, lower values are better and are indicated by $\downarrow$. For performance and combined metrics, higher values are better and are indicated by $\uparrow$. Bold values indicate the best result in each column for each model separately. \textbf{EQPO$_{ND}$} is the ablation of \textbf{EQPO} where the model does not have access to the ground truth demographic information, and the groups are inferred entirely via clustering. We release \textbf{MedGemma} trained with \textbf{EQPO} as \textbf{EquiMedGemma}. Results show mean $\pm$ std over 4 training runs. Detailed per dataset metrics are included in App. Tab. \ref{tab:detailed_demographics_repp_qwen}-\ref{tab:detailed_demographics_eqpo_medgemma}. }
\label{tab:fairness_performance_with_std}
\vspace{-0.5em}

\resizebox{\textwidth}{!}{
\small
\setlength{\tabcolsep}{4pt}
\begin{tabular}{l|ccccccc}
\toprule
\multirow{2}{*}{\textbf{Training Method}} & \multicolumn{7}{c}{\textbf{Fairness Metrics}} \\
\cmidrule(lr){2-8}
& \textbf{PP $\downarrow$} & \textbf{EOD $\downarrow$} & \textbf{FPR$_{\text{Diff}}$ $\downarrow$} & \textbf{$\sigma_{\text{F1}}$ $\downarrow$} & \textbf{$\Delta$F1 $\downarrow$} & \textbf{$\sigma_{\text{Acc}}$ $\downarrow$} & \textbf{$\Delta$Acc $\downarrow$} \\
\midrule
\multicolumn{8}{c}{\textbf{Qwen-2.5-VL-7B}} \\
\midrule
\textbf{Re++~\citep{hu2025reinforce}} & 16.66 $\pm$ 2.11 & 6.66 $\pm$ 1.59 & 6.37 $\pm$ 0.20 & .0322 $\pm$ .0000 & .0647 $\pm$ .0004 & 5.06 $\pm$ 0.49 & 10.33 $\pm$ 1.01 \\
\textbf{RLOO~\citep{rloo}} & 22.34 $\pm$ 0.86 & 6.67 $\pm$ 0.13 & 5.68 $\pm$ 0.80 & .0330 $\pm$ .0006 & .0693 $\pm$ .0017 & 4.86 $\pm$ 0.33 & 10.00 $\pm$ 0.79 \\
\textbf{GRPO~\citep{shao2024deepseekmath}} & 17.90 $\pm$ 9.21 & 7.93 $\pm$ 1.64 & \textbf{4.85 $\pm$ 0.34} & .0387 $\pm$ .0107 & .0821 $\pm$ .0215 & 4.85 $\pm$ 0.24 & 9.92 $\pm$ 0.69 \\
\textbf{GRPO+RS~\citep{puyol2021fairness}} & 19.62 $\pm$ 7.22 & 6.85 $\pm$ 0.80 & 6.44 $\pm$ 1.39 & .0319 $\pm$ .0009 & .0628 $\pm$ .0037 & 5.50 $\pm$ 0.17 & 11.26 $\pm$ 0.34 \\
\midrule
\textbf{EQPO} & \textbf{15.42 $\pm$ 1.95} & \textbf{5.62 $\pm$ 0.10} & 5.00 $\pm$ 0.87 & \textbf{.0254 $\pm$ .0035} & \textbf{.0522 $\pm$ .0099} & \textbf{4.42 $\pm$ 0.01} & \textbf{8.95 $\pm$ 0.03} \\
\bottomrule
\end{tabular}
}

\vspace{0.5em}

\resizebox{0.7\textwidth}{!}{
\small
\setlength{\tabcolsep}{4pt}
\begin{tabular}{l|cc|cc}
\toprule
\multirow{2}{*}{\textbf{Training Method}} & \multicolumn{2}{c|}{\textbf{Perf. Metrics}} & \multicolumn{2}{c}{\textbf{Combined}} \\
\cmidrule(lr){2-3} \cmidrule(lr){4-5}
& \textbf{Acc $\uparrow$} & \textbf{F1 $\uparrow$} & \textbf{Acc$_{\text{ES}}$ $\uparrow$} & \textbf{F1$_{\text{ES}}$ $\uparrow$} \\
\midrule
\multicolumn{5}{c}{\textbf{Qwen-2.5-VL-7B}} \\
\midrule
\textbf{Re++~\citep{hu2025reinforce}} & 75.31 $\pm$ 1.82 & .2599 $\pm$ .0065 & 71.69 $\pm$ 1.39 & .2518 $\pm$ .0063 \\
\textbf{RLOO~\citep{rloo}} & 78.22 $\pm$ 0.06 & .2523 $\pm$ .0013 & 74.59 $\pm$ 0.18 & .2443 $\pm$ .0014 \\
\textbf{GRPO~\citep{shao2024deepseekmath}} & 78.40 $\pm$ 0.69 & .2601 $\pm$ .0131 & 76.21 $\pm$ 0.91 & .2425 $\pm$ .0017 \\
\textbf{GRPO+RS~\citep{puyol2021fairness}} & 75.61 $\pm$ 2.96 & .2580 $\pm$ .0021 & 71.67 $\pm$ 2.92 & .2500 $\pm$ .0018 \\
\midrule
\textbf{EQPO} & \textbf{78.52 $\pm$ 0.31} & \textbf{.2657 $\pm$ .0036} & \textbf{77.14 $\pm$ 0.29} & \textbf{.2602 $\pm$ .0020} \\
\midrule
\multicolumn{5}{c}{\textbf{MedGemma-4B}} \\
\midrule
\textbf{Re++~\citep{hu2025reinforce}} & 78.76 $\pm$ 0.22 & .3105 $\pm$ .0179 & 75.41 $\pm$ 0.09 & .2966 $\pm$ .0191 \\
\textbf{RLOO~\citep{rloo}} & 79.76 $\pm$ 0.16 & .3237 $\pm$ .0019 & 76.48 $\pm$ 0.20 & .3056 $\pm$ .0021 \\
\textbf{GRPO~\citep{shao2024deepseekmath}} & 79.38 $\pm$ 0.15 & .3134 $\pm$ .0118 & 76.19 $\pm$ 0.12 & .3017 $\pm$ .0101 \\
\textbf{GRPO+RS~\citep{puyol2021fairness}} & 79.02 $\pm$ 0.15 & .2825 $\pm$ .0052 & 75.84 $\pm$ 0.30 & .2711 $\pm$ .0046 \\
\textbf{GRPO+DRO~\citep{sagawa2019distributionally}} & 80.17 $\pm$ 0.31 & .3146 $\pm$ .0177 & 76.69 $\pm$ 0.48 & .3009 $\pm$ .0173 \\
\midrule
\textbf{EQPO$_{ND}$} & 78.82 $\pm$ 0.58 & \textbf{.3484 $\pm$ .0041} & 75.67 $\pm$ 1.14 & \textbf{.3323 $\pm$ .0011} \\
\textbf{EQPO (EquiMedGemma)} & \textbf{80.40 $\pm$ 0.03} & .3275 $\pm$ .0007 & \textbf{77.23 $\pm$ 0.01} & .3155 $\pm$ .0006 \\
\bottomrule
\end{tabular}
}
\vspace{-1.5em}
\end{table*}

\begin{table*}[t]
\centering
\caption{\textbf{RQ1: Fairness and performance metrics for CheXpert dataset.} For fairness metrics, lower values are better and are indicated by $\downarrow$. For performance and combined metrics, higher values are better and are indicated by $\uparrow$. Bold values indicate the best result in each column.}
\label{tab:fairness_chexpert}
\vspace{-0.5em}
\resizebox{\textwidth}{!}{
\small
\setlength{\tabcolsep}{3pt}
\begin{tabular}{l|ccccccc|cc|cc}
\toprule
\multirow{2}{*}{\textbf{Training Method}} & \multicolumn{7}{c|}{\textbf{Fairness Metrics}} & \multicolumn{2}{c|}{\textbf{Perf. Metrics}} & \multicolumn{2}{c}{\textbf{Combined}} \\
\cmidrule(lr){2-8} \cmidrule(lr){9-10} \cmidrule(lr){11-12}
& \textbf{PP $\downarrow$} & \textbf{EOD $\downarrow$} & \textbf{FPR$_{\text{Diff}}$ $\downarrow$} & \textbf{$\sigma_{\text{F1}}$ $\downarrow$} & \textbf{$\Delta$F1 $\downarrow$} & \textbf{$\sigma_{\text{Acc}}$ $\downarrow$} & \textbf{$\Delta$Acc $\downarrow$} & \textbf{Acc $\uparrow$} & \textbf{F1 $\uparrow$} & \textbf{Acc$_{\text{ES}}$ $\uparrow$} & \textbf{F1$_{\text{ES}}$ $\uparrow$} \\
\midrule
\multicolumn{12}{c}{\textbf{Qwen-2.5-VL-7B}} \\
\midrule
\textbf{Re++} & $5.13_{\pm 2.55}$ & $3.74_{\pm 1.47}$ & $1.18_{\pm 0.35}$ & $.0148_{\pm .0082}$ & $.0282_{\pm .0142}$ & $5.86_{\pm 0.36}$ & $11.94_{\pm 0.72}$ & $77.30_{\pm 0.39}$ & $.1149_{\pm .0071}$ & $75.18_{\pm 0.35}$ & $.1257_{\pm .0095}$ \\
\textbf{RLOO} & $20.32_{\pm 7.47}$ & $6.25_{\pm 4.87}$ & $4.09_{\pm 1.37}$ & $.0259_{\pm .0164}$ & $.0529_{\pm .0388}$ & $5.42_{\pm 0.07}$ & $10.90_{\pm 0.07}$ & $77.74_{\pm 0.73}$ & $.1467_{\pm .0100}$ & $75.67_{\pm 0.57}$ & $.1621_{\pm .0215}$ \\
\textbf{GRPO} & $12.99_{\pm 1.45}$ & $4.49_{\pm 0.08}$ & $4.97_{\pm 0.03}$ & $.0194_{\pm .0042}$ & $.0386_{\pm .0028}$ & $5.23_{\pm 1.11}$ & $10.25_{\pm 2.60}$ & $77.79_{\pm 0.65}$ & $.1443_{\pm .0093}$ & $75.56_{\pm 0.87}$ & $.1572_{\pm .0074}$ \\
\textbf{GRPO+DRO} & $11.31_{\pm 2.86}$ & $3.91_{\pm 1.31}$ & $1.80_{\pm 0.09}$ & $.0177_{\pm .0034}$ & $.0339_{\pm .0034}$ & $6.38_{\pm 0.31}$ & $13.45_{\pm 0.67}$ & $76.91_{\pm 0.41}$ & $.1052_{\pm .0042}$ & $74.95_{\pm 0.50}$ & $.1168_{\pm .0061}$ \\
\textbf{EQPO} & $6.10_{\pm 1.94}$ & $7.80_{\pm 3.79}$ & $3.96_{\pm 0.19}$ & $.0234_{\pm .0051}$ & $.0439_{\pm .0152}$ & $4.99_{\pm 0.07}$ & $9.60_{\pm 0.28}$ & $78.62_{\pm 0.27}$ & $.1372_{\pm .0093}$ & $76.27_{\pm 0.29}$ & $.1510_{\pm .0110}$ \\
\bottomrule
\end{tabular}
}
\vspace{-1.5em}
\end{table*}

\begin{table*}[t]
\centering
\caption{\textbf{RQ1: Fairness and performance metrics for HAM10000 dataset.} For fairness metrics, lower values are better and are indicated by $\downarrow$. For performance and combined metrics, higher values are better and are indicated by $\uparrow$. Bold values indicate the best result in each column.}
\label{tab:fairness_ham10000}
\vspace{-0.5em}
\resizebox{\textwidth}{!}{
\small
\setlength{\tabcolsep}{3pt}
\begin{tabular}{l|ccccccc|cc|cc}
\toprule
\multirow{2}{*}{\textbf{Training Method}} & \multicolumn{7}{c|}{\textbf{Fairness Metrics}} & \multicolumn{2}{c|}{\textbf{Perf. Metrics}} & \multicolumn{2}{c}{\textbf{Combined}} \\
\cmidrule(lr){2-8} \cmidrule(lr){9-10} \cmidrule(lr){11-12}
& \textbf{PP $\downarrow$} & \textbf{EOD $\downarrow$} & \textbf{FPR$_{\text{Diff}}$ $\downarrow$} & \textbf{$\sigma_{\text{F1}}$ $\downarrow$} & \textbf{$\Delta$F1 $\downarrow$} & \textbf{$\sigma_{\text{Acc}}$ $\downarrow$} & \textbf{$\Delta$Acc $\downarrow$} & \textbf{Acc $\uparrow$} & \textbf{F1 $\uparrow$} & \textbf{Acc$_{\text{ES}}$ $\uparrow$} & \textbf{F1$_{\text{ES}}$ $\uparrow$} \\
\midrule
\multicolumn{12}{c}{\textbf{Qwen-2.5-VL-7B}} \\
\midrule
\textbf{Re++} & $22.90_{\pm 0.28}$ & $8.56_{\pm 3.49}$ & $4.72_{\pm 1.63}$ & $.0505_{\pm .0048}$ & $.1135_{\pm .0009}$ & $5.97_{\pm 1.49}$ & $12.28_{\pm 2.31}$ & $84.59_{\pm 2.76}$ & $.2167_{\pm .0216}$ & $83.21_{\pm 3.06}$ & $.2178_{\pm .0150}$ \\
\textbf{RLOO} & $44.69_{\pm 10.42}$ & $5.64_{\pm 0.08}$ & $5.81_{\pm 0.01}$ & $.0469_{\pm .0020}$ & $.1054_{\pm .0045}$ & $6.96_{\pm 0.13}$ & $13.77_{\pm 0.01}$ & $86.57_{\pm 0.04}$ & $.1974_{\pm .0000}$ & $85.43_{\pm 0.07}$ & $.2022_{\pm .0012}$ \\
\textbf{GRPO} & $35.65_{\pm 1.72}$ & $7.25_{\pm 0.13}$ & $7.08_{\pm 0.09}$ & $.0514_{\pm .0002}$ & $.1149_{\pm .0013}$ & $6.87_{\pm 0.06}$ & $13.75_{\pm 0.12}$ & $86.55_{\pm 0.10}$ & $.1860_{\pm .0137}$ & $85.36_{\pm 0.12}$ & $.1862_{\pm .0094}$ \\
\textbf{GRPO+DRO} & $36.69_{\pm 9.50}$ & $8.37_{\pm 2.25}$ & $6.11_{\pm 1.63}$ & $.0471_{\pm .0041}$ & $.0999_{\pm .0002}$ & $5.58_{\pm 1.90}$ & $11.03_{\pm 3.71}$ & $83.54_{\pm 4.18}$ & $.2140_{\pm .0253}$ & $82.49_{\pm 4.08}$ & $.2159_{\pm .0256}$ \\
\textbf{EQPO} & $26.56_{\pm 8.25}$ & $8.21_{\pm 1.08}$ & $5.98_{\pm 3.16}$ & $.0325_{\pm .0132}$ & $.0639_{\pm .0216}$ & $5.92_{\pm 0.42}$ & $11.64_{\pm 1.12}$ & $86.67_{\pm 0.64}$ & $.2738_{\pm .0006}$ & $85.71_{\pm 0.44}$ & $.2581_{\pm .0061}$ \\
\bottomrule
\end{tabular}
}
\vspace{-1.5em}
\end{table*}

\begin{table*}[t]
\centering
\caption{\textbf{RQ1: Fairness and performance metrics for ISIC2020 dataset.} For fairness metrics, lower values are better and are indicated by $\downarrow$. For performance and combined metrics, higher values are better and are indicated by $\uparrow$. Bold values indicate the best result in each column.}
\label{tab:fairness_isic2020}
\vspace{-0.5em}
\resizebox{\textwidth}{!}{
\small
\setlength{\tabcolsep}{3pt}
\begin{tabular}{l|ccccccc|cc|cc}
\toprule
\multirow{2}{*}{\textbf{Training Method}} & \multicolumn{7}{c|}{\textbf{Fairness Metrics}} & \multicolumn{2}{c|}{\textbf{Perf. Metrics}} & \multicolumn{2}{c}{\textbf{Combined}} \\
\cmidrule(lr){2-8} \cmidrule(lr){9-10} \cmidrule(lr){11-12}
& \textbf{PP $\downarrow$} & \textbf{EOD $\downarrow$} & \textbf{FPR$_{\text{Diff}}$ $\downarrow$} & \textbf{$\sigma_{\text{F1}}$ $\downarrow$} & \textbf{$\Delta$F1 $\downarrow$} & \textbf{$\sigma_{\text{Acc}}$ $\downarrow$} & \textbf{$\Delta$Acc $\downarrow$} & \textbf{Acc $\uparrow$} & \textbf{F1 $\uparrow$} & \textbf{Acc$_{\text{ES}}$ $\uparrow$} & \textbf{F1$_{\text{ES}}$ $\uparrow$} \\
\midrule
\multicolumn{12}{c}{\textbf{Qwen-2.5-VL-7B}} \\
\midrule
\textbf{Re++} & $21.51_{\pm 23.07}$ & $3.10_{\pm 3.06}$ & $3.06_{\pm 3.00}$ & $.0196_{\pm .0068}$ & $.0412_{\pm .0165}$ & $3.04_{\pm 0.84}$ & $6.56_{\pm 2.03}$ & $96.70_{\pm 2.16}$ & $.5205_{\pm .0212}$ & $95.51_{\pm 2.30}$ & $.5183_{\pm .0267}$ \\
\textbf{RLOO} & $2.59_{\pm 0.02}$ & $0.58_{\pm 0.47}$ & $0.47_{\pm 0.67}$ & $.0110_{\pm .0054}$ & $.0222_{\pm .0103}$ & $2.51_{\pm 0.08}$ & $5.26_{\pm 0.17}$ & $98.20_{\pm 0.03}$ & $.5004_{\pm .0073}$ & $97.08_{\pm 0.07}$ & $.4957_{\pm .0052}$ \\
\textbf{GRPO} & $26.52_{\pm 33.86}$ & $0.01_{\pm 0.02}$ & $0.01_{\pm 0.02}$ & $.0066_{\pm .0000}$ & $.0137_{\pm .0000}$ & $2.47_{\pm 0.01}$ & $5.17_{\pm 0.01}$ & $98.23_{\pm 0.00}$ & $.4955_{\pm .0000}$ & $97.14_{\pm 0.00}$ & $.4926_{\pm .0000}$ \\
\textbf{GRPO+DRO} & $14.75_{\pm 15.01}$ & $5.84_{\pm 7.91}$ & $5.83_{\pm 7.92}$ & $.0155_{\pm .0126}$ & $.0340_{\pm .0288}$ & $3.29_{\pm 1.13}$ & $7.00_{\pm 2.67}$ & $94.78_{\pm 4.26}$ & $.5056_{\pm .0159}$ & $93.69_{\pm 4.42}$ & $.5053_{\pm .0192}$ \\
\textbf{EQPO} & $2.59_{\pm 0.02}$ & $0.12_{\pm 0.17}$ & $0.00_{\pm 0.00}$ & $.0068_{\pm .0003}$ & $.0141_{\pm .0006}$ & $2.50_{\pm 0.05}$ & $5.23_{\pm 0.10}$ & $98.14_{\pm 0.14}$ & $.4950_{\pm .0007}$ & $97.04_{\pm 0.15}$ & $.4921_{\pm .0008}$ \\
\bottomrule
\end{tabular}
}
\vspace{-1.5em}
\end{table*}

\begin{table*}[t]
\centering
\caption{\textbf{RQ1: Fairness and performance metrics for PAD-UFES-20 dataset.} For fairness metrics, lower values are better and are indicated by $\downarrow$. For performance and combined metrics, higher values are better and are indicated by $\uparrow$. Bold values indicate the best result in each column.}
\label{tab:fairness_padufes}
\vspace{-0.5em}
\resizebox{\textwidth}{!}{
\small
\setlength{\tabcolsep}{3pt}
\begin{tabular}{l|ccccccc|cc|cc}
\toprule
\multirow{2}{*}{\textbf{Training Method}} & \multicolumn{7}{c|}{\textbf{Fairness Metrics}} & \multicolumn{2}{c|}{\textbf{Perf. Metrics}} & \multicolumn{2}{c}{\textbf{Combined}} \\
\cmidrule(lr){2-8} \cmidrule(lr){9-10} \cmidrule(lr){11-12}
& \textbf{PP $\downarrow$} & \textbf{EOD $\downarrow$} & \textbf{FPR$_{\text{Diff}}$ $\downarrow$} & \textbf{$\sigma_{\text{F1}}$ $\downarrow$} & \textbf{$\Delta$F1 $\downarrow$} & \textbf{$\sigma_{\text{Acc}}$ $\downarrow$} & \textbf{$\Delta$Acc $\downarrow$} & \textbf{Acc $\uparrow$} & \textbf{F1 $\uparrow$} & \textbf{Acc$_{\text{ES}}$ $\uparrow$} & \textbf{F1$_{\text{ES}}$ $\uparrow$} \\
\midrule
\multicolumn{12}{c}{\textbf{Qwen-2.5-VL-7B}} \\
\midrule
\textbf{Re++} & $34.76_{\pm 1.25}$ & $13.01_{\pm 5.64}$ & $10.14_{\pm 0.71}$ & $.0589_{\pm .0255}$ & $.1231_{\pm .0543}$ & $3.73_{\pm 0.78}$ & $7.58_{\pm 1.99}$ & $77.96_{\pm 0.07}$ & $.3129_{\pm .0035}$ & $79.42_{\pm 0.20}$ & $.3121_{\pm .0122}$ \\
\textbf{RLOO} & $36.69_{\pm 0.65}$ & $16.74_{\pm 1.42}$ & $10.12_{\pm 0.27}$ & $.0851_{\pm .0019}$ & $.1788_{\pm .0034}$ & $6.15_{\pm 0.27}$ & $12.13_{\pm 0.09}$ & $77.12_{\pm 0.37}$ & $.2672_{\pm .0078}$ & $79.97_{\pm 0.31}$ & $.2788_{\pm .0081}$ \\
\textbf{GRPO} & $37.46_{\pm 6.33}$ & $16.50_{\pm 4.29}$ & $13.89_{\pm 6.40}$ & $.0826_{\pm .0120}$ & $.1686_{\pm .0258}$ & $4.90_{\pm 1.22}$ & $10.18_{\pm 2.89}$ & $76.70_{\pm 0.67}$ & $.2614_{\pm .0309}$ & $78.72_{\pm 0.96}$ & $.2684_{\pm .0241}$ \\
\textbf{GRPO+DRO} & $37.38_{\pm 2.46}$ & $15.56_{\pm 2.88}$ & $9.36_{\pm 1.06}$ & $.0680_{\pm .0209}$ & $.1346_{\pm .0330}$ & $4.20_{\pm 0.03}$ & $8.74_{\pm 0.22}$ & $78.22_{\pm 0.30}$ & $.3271_{\pm .0088}$ & $79.94_{\pm 0.69}$ & $.3229_{\pm .0142}$ \\
\textbf{EQPO} & $24.50_{\pm 9.47}$ & $8.26_{\pm 3.84}$ & $10.14_{\pm 1.15}$ & $.0516_{\pm .0031}$ & $.1075_{\pm .0115}$ & $1.92_{\pm 0.30}$ & $4.09_{\pm 0.46}$ & $80.31_{\pm 2.52}$ & $.2995_{\pm .0547}$ & $78.70_{\pm 2.74}$ & $.2923_{\pm .0486}$ \\
\bottomrule
\end{tabular}
}
\vspace{-1.5em}
\end{table*}

\begin{table*}[t]
\centering
\caption{\textbf{RQ1: Fairness and performance metrics for Hemorrhage dataset.} For fairness metrics, lower values are better and are indicated by $\downarrow$. For performance and combined metrics, higher values are better and are indicated by $\uparrow$. Bold values indicate the best result in each column.}
\label{tab:fairness_hemorrhage}
\vspace{-0.5em}
\resizebox{\textwidth}{!}{
\small
\setlength{\tabcolsep}{3pt}
\begin{tabular}{l|ccccccc|cc|cc}
\toprule
\multirow{2}{*}{\textbf{Training Method}} & \multicolumn{7}{c|}{\textbf{Fairness Metrics}} & \multicolumn{2}{c|}{\textbf{Perf. Metrics}} & \multicolumn{2}{c}{\textbf{Combined}} \\
\cmidrule(lr){2-8} \cmidrule(lr){9-10} \cmidrule(lr){11-12}
& \textbf{PP $\downarrow$} & \textbf{EOD $\downarrow$} & \textbf{FPR$_{\text{Diff}}$ $\downarrow$} & \textbf{$\sigma_{\text{F1}}$ $\downarrow$} & \textbf{$\Delta$F1 $\downarrow$} & \textbf{$\sigma_{\text{Acc}}$ $\downarrow$} & \textbf{$\Delta$Acc $\downarrow$} & \textbf{Acc $\uparrow$} & \textbf{F1 $\uparrow$} & \textbf{Acc$_{\text{ES}}$ $\uparrow$} & \textbf{F1$_{\text{ES}}$ $\uparrow$} \\
\midrule
\multicolumn{12}{c}{\textbf{Qwen-2.5-VL-7B}} \\
\midrule
\textbf{Re++} & $14.94_{\pm 0.12}$ & $9.35_{\pm 5.58}$ & $9.35_{\pm 5.58}$ & $.0557_{\pm .0345}$ & $.0890_{\pm .0574}$ & $4.20_{\pm 0.42}$ & $7.29_{\pm 0.48}$ & $75.83_{\pm 3.71}$ & $.4822_{\pm .0246}$ & $75.28_{\pm 2.94}$ & $.4926_{\pm .0332}$ \\
\textbf{RLOO} & $13.23_{\pm 10.27}$ & $1.74_{\pm 0.41}$ & $1.74_{\pm 0.41}$ & $.0213_{\pm .0049}$ & $.0397_{\pm .0079}$ & $6.51_{\pm 1.39}$ & $12.22_{\pm 2.25}$ & $81.26_{\pm 0.41}$ & $.4483_{\pm .0013}$ & $79.53_{\pm 0.48}$ & $.4420_{\pm .0018}$ \\
\textbf{GRPO} & $12.58_{\pm 7.06}$ & $3.74_{\pm 1.44}$ & $3.74_{\pm 1.44}$ & $.0302_{\pm .0269}$ & $.0495_{\pm .0455}$ & $5.48_{\pm 0.97}$ & $9.90_{\pm 1.27}$ & $80.78_{\pm 2.75}$ & $.4805_{\pm .0314}$ & $79.28_{\pm 2.79}$ & $.4811_{\pm .0349}$ \\
\textbf{GRPO+DRO} & $16.38_{\pm 5.61}$ & $5.96_{\pm 1.93}$ & $5.96_{\pm 1.93}$ & $.0407_{\pm .0056}$ & $.0639_{\pm .0146}$ & $6.63_{\pm 1.20}$ & $12.12_{\pm 1.93}$ & $79.71_{\pm 2.06}$ & $.4886_{\pm .0167}$ & $78.52_{\pm 1.91}$ & $.4923_{\pm .0160}$ \\
\textbf{EQPO} & $34.02_{\pm 21.84}$ & $1.59_{\pm 2.00}$ & $1.59_{\pm 2.00}$ & $.0263_{\pm .0096}$ & $.0501_{\pm .0212}$ & $6.14_{\pm 0.09}$ & $10.91_{\pm 0.08}$ & $83.30_{\pm 0.00}$ & $.5011_{\pm .0660}$ & $81.84_{\pm 0.30}$ & $.4970_{\pm .0685}$ \\
\bottomrule
\end{tabular}
}
\vspace{-1.5em}
\end{table*}

\begin{table*}[t]
\centering
\caption{\textbf{RQ1: Fairness and performance metrics for VinDr dataset.} For fairness metrics, lower values are better and are indicated by $\downarrow$. For performance and combined metrics, higher values are better and are indicated by $\uparrow$. Bold values indicate the best result in each column.}
\label{tab:fairness_vindr}
\vspace{-0.5em}
\resizebox{\textwidth}{!}{
\small
\setlength{\tabcolsep}{3pt}
\begin{tabular}{l|ccccccc|cc|cc}
\toprule
\multirow{2}{*}{\textbf{Training Method}} & \multicolumn{7}{c|}{\textbf{Fairness Metrics}} & \multicolumn{2}{c|}{\textbf{Perf. Metrics}} & \multicolumn{2}{c}{\textbf{Combined}} \\
\cmidrule(lr){2-8} \cmidrule(lr){9-10} \cmidrule(lr){11-12}
& \textbf{PP $\downarrow$} & \textbf{EOD $\downarrow$} & \textbf{FPR$_{\text{Diff}}$ $\downarrow$} & \textbf{$\sigma_{\text{F1}}$ $\downarrow$} & \textbf{$\Delta$F1 $\downarrow$} & \textbf{$\sigma_{\text{Acc}}$ $\downarrow$} & \textbf{$\Delta$Acc $\downarrow$} & \textbf{Acc $\uparrow$} & \textbf{F1 $\uparrow$} & \textbf{Acc$_{\text{ES}}$ $\uparrow$} & \textbf{F1$_{\text{ES}}$ $\uparrow$} \\
\midrule
\multicolumn{12}{c}{\textbf{Qwen-2.5-VL-7B}} \\
\midrule
\textbf{Re++} & $17.41_{\pm 4.38}$ & $8.87_{\pm 6.14}$ & $16.16_{\pm 8.70}$ & $.0261_{\pm .0030}$ & $.0581_{\pm .0016}$ & $12.58_{\pm 2.80}$ & $26.64_{\pm 5.98}$ & $75.51_{\pm 6.11}$ & $.1676_{\pm .0447}$ & $68.59_{\pm 3.89}$ & $.1530_{\pm .0321}$ \\
\textbf{RLOO} & $29.55_{\pm 0.84}$ & $10.91_{\pm 3.42}$ & $15.38_{\pm 2.59}$ & $.0314_{\pm .0072}$ & $.0742_{\pm .0162}$ & $10.68_{\pm 1.82}$ & $23.76_{\pm 3.75}$ & $86.62_{\pm 0.16}$ & $.1704_{\pm .0051}$ & $79.25_{\pm 1.16}$ & $.1705_{\pm .0026}$ \\
\textbf{GRPO} & $31.91_{\pm 2.14}$ & $13.33_{\pm 0.00}$ & $12.50_{\pm 1.63}$ & $.0389_{\pm .0040}$ & $.0918_{\pm .0087}$ & $9.33_{\pm 0.09}$ & $20.89_{\pm 0.19}$ & $86.58_{\pm 0.21}$ & $.1689_{\pm .0030}$ & $80.01_{\pm 0.10}$ & $.1745_{\pm .0030}$ \\
\textbf{GRPO+DRO} & $20.81_{\pm 15.05}$ & $8.29_{\pm 2.30}$ & $16.00_{\pm 0.52}$ & $.0345_{\pm .0087}$ & $.0730_{\pm .0189}$ & $12.43_{\pm 2.80}$ & $26.47_{\pm 4.88}$ & $76.77_{\pm 8.94}$ & $.1625_{\pm .0524}$ & $69.69_{\pm 9.11}$ & $.1527_{\pm .0542}$ \\
\textbf{EQPO} & $14.19_{\pm 5.07}$ & $13.35_{\pm 0.00}$ & $13.36_{\pm 0.04}$ & $.0369_{\pm .0002}$ & $.0856_{\pm .0000}$ & $9.44_{\pm 0.01}$ & $21.20_{\pm 0.01}$ & $86.82_{\pm 0.02}$ & $.1608_{\pm .0004}$ & $80.12_{\pm 0.02}$ & $.1696_{\pm .0003}$ \\
\bottomrule
\end{tabular}
}
\vspace{-1.5em}
\end{table*}

\clearpage


\newpage
\section*{NeurIPS Paper Checklist}

\begin{enumerate}

\item {\bf Claims}
    \item[] Question: Do the main claims made in the abstract and introduction accurately reflect the paper's contributions and scope?
    \item[] Answer: \answerYes{}
    \item[] Justification: The abstract and contributions in §\ref{sec:eqpo} state three claims and each is supported in the cited section. The claim that EQPO reduces predictive parity by 27.2\% and improves F1 by 12.5\% over RL baselines is supported by Tab.~\ref{tab:fairness_performance} and §\ref{sec:experiment_setup}. The claim that reward-based clustering recovers demographically meaningful structure without supervision is supported by Fig.~\ref{fig:cluster_contingency} and §\ref{sec:cluster_stability}. The release of EquiMedGemma-4B is provided through the anonymous link at the end of the abstract.

\item {\bf Limitations}
    \item[] Question: Does the paper discuss the limitations of the work performed by the authors?
    \item[] Answer: \answerYes{}
    \item[] Justification: A dedicated Limitations section is included as App.~\ref{app:limitations}, covering the heuristic temperature design, the gap between predictive-parity metrics and clinical-outcome equity, possible bias amplification when inferred clusters correlate with confounded attributes, and the research-prototype scope (7 datasets across 5 imaging modalities, no time-series or EHR validation).

\item {\bf Theory assumptions and proofs}
    \item[] Question: For each theoretical result, does the paper provide the full set of assumptions and a complete (and correct) proof?
    \item[] Answer: \answerYes{}
    \item[] Justification: A proof sketch for the convergence claim appears in §\ref{sec:eqpo} under the paragraph ``Convergence Analysis'', and the full derivation is given in App.~\ref{app:theory}. The setup is stated explicitly: two groups with imbalanced counts, approximately equal initial rewards, the GRPO unit-variance normalization, and the bounded KL budget. The appendix covers both the $\beta = 0$ per-step gradient allocation and the finite-KL convergence argument.

    \item {\bf Experimental result reproducibility}
    \item[] Question: Does the paper fully disclose all the information needed to reproduce the main experimental results of the paper to the extent that it affects the main claims and/or conclusions of the paper (regardless of whether the code and data are provided or not)?
    \item[] Answer: \answerYes{}
    \item[] Justification: §\ref{sec:experiment_setup} specifies datasets, demographic group definitions, evaluation metrics, and baselines; App.~\ref{app:hyperparameters} lists every hyperparameter (batch sizes, learning rate, rollout count, GPU count, KL setting, etc.); App.~\ref{app:eval_metrics} gives the mathematical definitions of each fairness and performance metric; App.~\ref{app:pseudocode} contains pseudocode for both GRPO and EQPO advantage computation. Code is also released at the anonymous link in the abstract.

\item {\bf Open access to data and code}
    \item[] Question: Does the paper provide open access to the data and code, with sufficient instructions to faithfully reproduce the main experimental results, as described in supplemental material?
    \item[] Answer: \answerYes{}
    \item[] Justification: All seven evaluation datasets are publicly available and cited in §\ref{sec:experiment_setup}; our training pipeline, evaluation framework, and the EquiMedGemma-4B checkpoint are released at the anonymous link given in the abstract.

\item {\bf Experimental setting/details}
    \item[] Question: Does the paper specify all the training and test details (e.g., data splits, hyperparameters, how they were chosen, type of optimizer) necessary to understand the results?
    \item[] Answer: \answerYes{}
    \item[] Justification: §\ref{sec:experiment_setup} describes the training protocol (unified 15-epoch finetuning across all 7 datasets), demographic group definitions (gender; four 25-year age brackets), and the hierarchical-averaging evaluation strategy. App.~\ref{app:hyperparameters} provides the full hyperparameter table (learning rate $5\times10^{-7}$, train/val batch size 512, 10 rollouts per prompt, KL disabled, vLLM rollout engine, 4 H200 GPUs).

\item {\bf Experiment statistical significance}
    \item[] Question: Does the paper report error bars suitably and correctly defined or other appropriate information about the statistical significance of the experiments?
    \item[] Answer: \answerYes{}
    \item[] Justification: All main-table results are averaged over four independent training runs, as stated in the caption of Tab.~\ref{tab:fairness_performance}. The corresponding mean $\pm$ standard deviation values are reported in App. Tab.~\ref{tab:fairness_performance_with_std}.

\item {\bf Experiments compute resources}
    \item[] Question: For each experiment, does the paper provide sufficient information on the computer resources (type of compute workers, memory, time of execution) needed to reproduce the experiments?
    \item[] Answer: \answerYes{}
    \item[] Justification: §\ref{sec:experiment_setup} states that all experiments use 4 NVIDIA H200 GPUs on a single node. App.~\ref{app:hyperparameters} adds the per-run details (tensor model parallel size 2, GPU memory utilization 0.6 for vLLM rollouts, 15 epochs) sufficient to estimate wall-clock cost.

\item {\bf Code of ethics}
    \item[] Question: Does the research conducted in the paper conform, in every respect, with the NeurIPS Code of Ethics \url{https://neurips.cc/public/EthicsGuidelines}?
    \item[] Answer: \answerYes{}
    \item[] Justification: The research was conducted in accordance with the NeurIPS Code of Ethics. All datasets are publicly available and pre-anonymized; no new human subject data was collected; the released model is documented with intended-use and limitations statements; and demographic fairness is the primary objective of the work.

\item {\bf Broader impacts}
    \item[] Question: Does the paper discuss both potential positive societal impacts and negative societal impacts of the work performed?
    \item[] Answer: \answerYes{}
    \item[] Justification: A dedicated ``Impact Statement'' section in §6 discusses the positive impact (reducing demographic disparities in clinical AI to support more equitable diagnostic outcomes) and the negative considerations (incomplete coverage of patient diversity dimensions, the requirement for regulatory approval before clinical deployment, and reliance on the underlying license terms of the source datasets).

\item {\bf Safeguards}
    \item[] Question: Does the paper describe safeguards that have been put in place for responsible release of data or models that have a high risk for misuse (e.g., pre-trained language models, image generators, or scraped datasets)?
    \item[] Answer: \answerNA{}
    \item[] Justification: EquiMedGemma-4B is a fairness-tuned diagnostic model derived from MedGemma-4B-IT and inherits its base license; it is intended for research use only, as stated in the Impact Statement, and does not present a misuse profile beyond that of standard medical VLLMs. No high-risk generative content (e.g., deepfakes, scraped private images) is released.

\item {\bf Licenses for existing assets}
    \item[] Question: Are the creators or original owners of assets (e.g., code, data, models), used in the paper, properly credited and are the license and terms of use explicitly mentioned and properly respected?
    \item[] Answer: \answerYes{}
    \item[] Justification: Every dataset (CheXpert, COVID-BLUES, VinDr-Mammo, ISIC-2020, HAM10000, PAD-UFES-20, Hemorrhage), every base model (Qwen-2.5-VL, MedGemma, QoQ-Med3), and every framework (VERL, vLLM) is cited in §\ref{sec:experiment_setup} or App.~\ref{app:hyperparameters}. All datasets are publicly available under their respective licenses, and the base-model licenses are inherited by their fine-tuned counterparts.

\item {\bf New assets}
    \item[] Question: Are new assets introduced in the paper well documented and is the documentation provided alongside the assets?
    \item[] Answer: \answerYes{}
    \item[] Justification: The released EquiMedGemma-4B checkpoint, training pipeline, and evaluation framework are accompanied by setup instructions, hyperparameter configurations, and an intended-use statement at the anonymous repository linked in the abstract. The repository is anonymized for double-blind review.

\item {\bf Crowdsourcing and research with human subjects}
    \item[] Question: For crowdsourcing experiments and research with human subjects, does the paper include the full text of instructions given to participants and screenshots, if applicable, as well as details about compensation (if any)?
    \item[] Answer: \answerNA{}
    \item[] Justification: This work involves no crowdsourcing and no new research with human subjects; all clinical data is reused from publicly released, pre-anonymized datasets.

\item {\bf Institutional review board (IRB) approvals or equivalent for research with human subjects}
    \item[] Question: Does the paper describe potential risks incurred by study participants, whether such risks were disclosed to the subjects, and whether Institutional Review Board (IRB) approvals (or an equivalent approval/review based on the requirements of your country or institution) were obtained?
    \item[] Answer: \answerNA{}
    \item[] Justification: No new human subjects research was conducted. All data is reused from publicly available, pre-anonymized clinical datasets that obtained their own ethical approvals at the time of original release.

\item {\bf Declaration of LLM usage}
    \item[] Question: Does the paper describe the usage of LLMs if it is an important, original, or non-standard component of the core methods in this research? Note that if the LLM is used only for writing, editing, or formatting purposes and does \emph{not} impact the core methodology, scientific rigor, or originality of the research, declaration is not required.
    \item[] Answer: \answerYes{}
    \item[] Justification: The trained subjects of our work are themselves vision-LLMs (Qwen-2.5-VL-7B, MedGemma-4B, QoQ-Med3-8B), cited in §\ref{sec:experiment_setup}. In addition, an LLM-as-a-judge auxiliary experiment using Gemini-3.1-Flash-Lite is reported in App.~\ref{app:judge_reward}, and ChatGPT was used solely for grammar correction and debugging assistance, as disclosed in the appendix's ``Use of Large Language Models'' section.

\end{enumerate}

\end{document}